\documentclass[journal]{IEEEtran}

\usepackage{graphics} % for pdf, bitmapped graphics files
\usepackage{epsfig} % for postscript graphics files
\usepackage{amsmath, amsfonts} % assumes amsmath package installed
\usepackage{amssymb}  % assumes amssymb package installed
\usepackage{amsthm}
\usepackage{algorithm}
\usepackage{algpseudocode}
\usepackage{tikz}
\usepackage[hidelinks]{hyperref}
\usepackage[capitalize]{cleveref}
\usepackage{mathnotation}
\usepackage{xcolor}
\usepackage{cite}

\newcommand{\dbstatus}{no}

\newcommand{\ifstringequal}[4]{%
  \ifnum\pdfstrcmp{#1}{#2}=0
  #3%
  \else
  #4%
  \fi
}

\newcommand{\db}[2]{%
  \ifstringequal{\dbstatus}{yes}
    {#1}
    {#2}
}

\AtBeginEnvironment{appendices}{\crefalias{section}{appendix}}

\begin{document}

\title{The Geometry of Optimal Gait Families for \\ Steering Kinematic Locomoting Systems}

\author{Jinwoo~Choi,~\IEEEmembership{Student Member,~IEEE,}
        Siming~Deng,~\IEEEmembership{Student Member,~IEEE,}
		Nathan~Justus,~\IEEEmembership{Student Member,~IEEE,}
		Noah~J.~Cowan,~\IEEEmembership{Fellow,~IEEE}
		and~Ross~L.~Hatton,~\IEEEmembership{Member,~IEEE}% <-this % stops a space
\thanks{This work was supported in part by the National Science Foundation under awards CMMI-1653220 and 1826446 and by the Office of Naval Research via Award N00014-23-1-2171}% <-this % stops a space
\thanks{
	Jinwoo Choi, Nathan Justus, and Ross L. Hatton are with the Collaborative Robotics and Intelligent Systems (CoRIS) Institute at Oregon State University, Corvallis, OR USA.}% <-this % stops a space
\thanks{
	Siming Deng and Noah J. Cowan are with the Laboratory for Computational Sensing and Robotics, and the Department of Mechanical Engineering, Johns Hopkins University, Baltimore MD 21218 USA.}% <-this % stops a space
}

% \markboth{Journal of \LaTeX\ Class Files,~Vol.~18, No.~9, September~2020}%
% {How to Use the IEEEtran \LaTeX \ Templates}

\theoremstyle{definition}
\newtheorem{theorem}{Theorem}[]

\theoremstyle{definition}
\newtheorem{definition}{Definition}[]

\maketitle

\newcommand{\grad}[1][]{\nabla_{#1}}

\newcommand{\gaitcost}[0]{\alnth_{\gait}}
\newcommand{\gaitdisp}[1][]{\fiber_\gait^{#1}}
\newcommand{\gaitexpdisp}[1][]{z_\gait^{#1}}
\newcommand{\gaitefffrac}[1][]{\frac{\gaitdisp{#1}}{\gaitcost}}
\newcommand{\dgaitefffrac}[1]{\grad[#1]\gaitdisp-\frac{\gaitdisp}{\gaitcost}\grad[#1]{\gaitcost}}

\newcommand{\eff}{\eta}
\newcommand{\gaiteff}[1][]{\eta_\gait^{#1}}
\newcommand{\gaiteffx}{\gaiteff[x]}
\newcommand{\gaitefftheta}{\gaiteff[\theta]}
\newcommand{\gaiteffxnorm}{\overline{\eta}_\gait^x}
\newcommand{\gaiteffthetanorm}{\overline{\eta}_\gait^\theta}

\newcommand{\dtensor}{D_r}
\newcommand{\mtensor}{M_r}
\newcommand{\dualmtensor}{\mtensor^*}
\newcommand{\comtensor}{\tilde{M}_r}
\newcommand{\coriolis}{C}
\newcommand{\gcurve}{\gamma}
\newcommand{\covacc}{a_{r}}

\newcommand{\rank}{\text{rank}}
\newcommand{\lagrfun}{\mathcal{L}}
\newcommand{\unconstmax}[2]{\underset{#1}{\text{maximize }} \ \ {#2}}
\newcommand{\unconstmin}[2]{\underset{#1}{\text{minimize }} \ \ {#2}}
\newcommand{\constoptone}[4]{
	\begin{split}
		\underset{#2}{#1} & \ {#3} \\
		\text{subject to } & \ {#4} 
	\end{split}
	}
\newcommand{\constopttwo}[5]{
	\begin{split}
		\underset{#2}{#1} & \ {#3} \\
		\text{subject to } & \ {#4} \\			
		& \ {#5} 
	\end{split}
	}

\newcommand{\constminone}[3]{\constoptone{\text{minimize }}{#1}{#2}{#3}}
\newcommand{\constmintwo}[4]{\constopttwo{\text{minimize }}{#1}{#2}{#3}{#4}}
\newcommand{\constmaxone}[3]{\constoptone{\text{maximize }}{#1}{#2}{#3}}
\newcommand{\constmaxtwo}[4]{\constopttwo{\text{maximize }}{#1}{#2}{#3}{#4}}

\newcommand{\opt}[1]{{#1}^{*}}
\newcommand{\gaitparam}{p}
\newcommand{\gaitparamspace}{\mathcal{P}}
\newcommand{\contpar}{c}
\newcommand{\contparspace}{\mathcal{C}}
\newcommand{\contparmin}{\contpar_{\min}}
\newcommand{\contparmax}{\contpar_{\max}}
\newcommand{\cimin}{c_{i,\min}}
\newcommand{\cimax}{c_{i,\max}}
\newcommand{\contfcn}{H}
\newcommand{\contvar}{q}
\newcommand{\contvarspace}{\mathcal{Q}}
\newcommand{\contsol}{\gamma}
\newcommand{\contsolspace}{\Gamma}
\newcommand{\contsoltanspace}[1][]{\tans[#1]\contsolspace}
\newcommand{\contpredstep}{\Delta\contsol_{\text{pred}}}
\newcommand{\contcorrstep}{\contsol_{\text{corr}}}
\newcommand{\contsearchvec}{e}
\newcommand{\contmidstep}{\upsilon}

\newcommand{\objfun}[1][]{f_{#1}}
\newcommand{\objmult}[1][]{\sigma_{#1}}
\newcommand{\eqfun}[1][]{h_{#1}}
\newcommand{\eqmult}[1][]{\lambda_{#1}}
\newcommand{\numeq}{m_1}
\newcommand{\ineqfun}[1][]{y_{#1}}
\newcommand{\ineqmult}[1][]{\mu_{#1}}
\newcommand{\numineq}{m_2}
\newcommand{\ncpfun}{\chi}
\newcommand{\ncpfunset}{X}

\newcommand{\forvel}{v}
\newcommand{\rotvel}{\omega}

\newcommand{\anchoreff}[1]{\eff_{\gait_{#1}}}
\newcommand{\paretorad}{\rho_\gait}
\newcommand{\paretoangle}{\varphi_\gait}
\newcommand{\nullcont}{\mathcal{N}_\contfcn}

%%%%%%%%%%%%%%%%%%%%%%%%%%%%%%%%%%%%%%%%%%%%%%%%%%%%%%%%%%%%%%%%%%%%%%%%%%%%%%%%
% As a general rule, do not put math, special symbols or citations
% in the abstract or keywords.
\begin{abstract}
Motion planning for locomotion systems typically requires translating high-level rigid-body tasks into low-level joint trajectories—a process that is straightforward for car-like robots with fixed, unbounded actuation inputs but more challenging for systems like snake robots, where the mapping depends on the current configuration and is constrained by joint limits. In this paper, we focus on generating continuous families of optimal gaits—collections of gaits parameterized by step size or steering rate—to enhance controllability and maneuverability. We uncover the underlying geometric structure of these optimal gait families and propose methods for constructing them using both global and local search strategies, where the local method and the global method compensate each other. The global search approach is robust to nonsmooth behavior, albeit yielding reduced-order solutions, while the local search provides higher accuracy but can be unstable near nonsmooth regions. To demonstrate our framework, we generate optimal gait families for viscous and perfect-fluid three-link swimmers. This work lays a foundation for integrating low-level joint controllers with higher-level motion planners in complex locomotion systems.
\end{abstract}
	
% Note that keywords are not normally used for peerreview papers.

% For peer review papers, you can put extra information on the cover
% page as needed:
% \ifCLASSOPTIONpeerreview
% \begin{center} \bfseries EDICS Category: 3-BBND \end{center}
% \fi
%
% For peerreview papers, this IEEEtran command inserts a page break and
% creates the second title. It will be ignored for other modes.
\IEEEpeerreviewmaketitle

%%%%%%%%%%%%%%%%%%%%%%%%%%%%%%%%%%%%%%%%%%%%%%%%%%%%%%%%%%%%%%%%%%%%%%%%%%%%%%%%
\section{Introduction}

\begin{figure*}
	\centering
	\includegraphics[width=\textwidth]{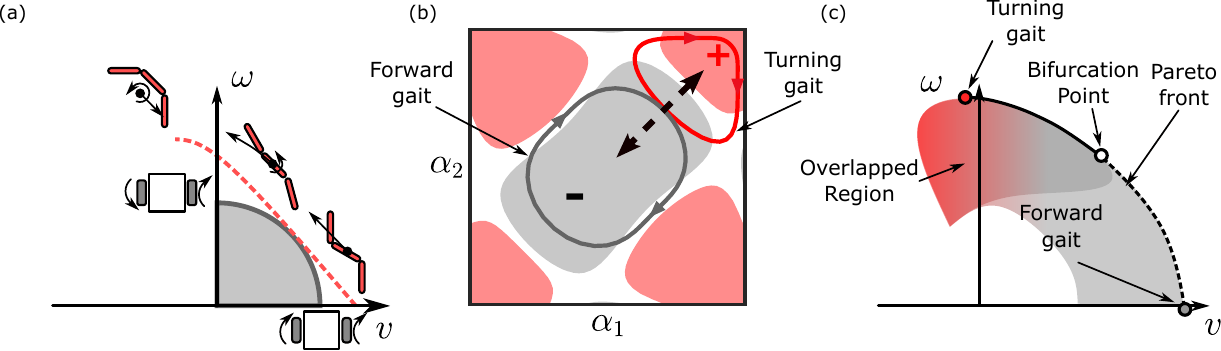}
	\caption{Schematic comparing the command spaces of a differential-drive car~\cite{laumond1998robot} and the Pareto optimal gaits (act as the boundary of the command spaces) of a viscous three-link swimmer~\cite{hatton2011geometric}. The horizontal axis represents forward velocity $v$, and the vertical axis represents angular velocity $\omega$. The gray and red curves respectively depict the boundaries of the command spaces for the differential-drive car and the viscous three-link swimmer, under unit power consumption. For the differential-drive car, this boundary is a circular arc. The arrows on the car indicate the direction of the corresponding wheel's rotation. The arrows on the swimmer indicate the corresponding body motion. Only forward-right-turning motions of viscous swimmers are shown. The velocity norm ${v^2+\omega^2}$ is defined as the control effort for the differential-drive car. (b) Schematic of \emph{forward function} for the viscous three-link swimmers. The grey and red regions denote the negative and positive areas of the height function, respectively. The grey and red arrow-marked curves are the optimal gait cycles in the $x$ and $\theta$ direction, respectively. The net displacement is the area integral enclosed by each gait. The forward gait encloses the grey region while minimizing the cost (or pathlength). (c) Schematic of an optimal gait family and its Pareto front~\cite{deng2022enhancing} for the viscous three-link swimmer in the average-velocity space. Each axis corresponds to the unit-effort averaged forward or angular velocity, $v$ and $\omega$, produced by each gait. The red dot indicates the turning gait, and the gray dot indicates the forward gait. The black solid and dashed curves connect these gaits on the Pareto front, with the solid portion denoting the front in an overlapping region of the command space.}
	\label{fig:GraphicalAbstract}
\end{figure*}

\IEEEPARstart{A}{} standard approach to motion planning for mobile robots is to treat the system as a rigid body equipped with a set of control fields defined in the body frame, and then construct a trajectory through $SE(2)$ or $SE(3)$ that can be achieved via the controls and which satisfies task constraints (e.g., avoiding obstacles and visiting regions of interest). For vehicle-type systems, this approach is facilitated by the system architecture: either the relationship between the control inputs and the rigid body motion is fixed and unbounded (as in the case of differential-drive cars or multi-rotor air- or watercraft) or the system has unbounded ``drive" actuators whose action on the rigid body is modulated via ``steering" actuators which do not directly propel the system (as in the case of Ackerman cars or air- and watercraft that combine a single propulsive thruster with rudders and other control surfaces).

Extending this approach to more general locomoting systems (e.g., snake robots, flap-wing, and swimming systems) introduces significant challenges. In particular, the direct control inputs for these systems combine drive and steering functions---changing shape pushes off against the environment, but the direction of induced motion depends on the shape the system is in. Additionally, shape changes are typically bounded by joint limits and other self-collision concerns, so it is not generally possible to simply ``turn on" a control mode in the same manner as for an unrestricted wheel, and the system inputs must instead oscillate the system joints within their limits.

These challenges can be addressed by adopting a hierarchical maneuver-based planning approach, in which the control fields from the vehicle scenario are replaced by a set of maneuvers---predefined shape oscillations (``gaits") that produce characteristic net displacements, such that sequences of these maneuvers can be used to on-average follow a specified rigid-body trajectory~\cite{frazzoli2001robust,frazz2005maneuverbased,niu2013locomotion,da20162d,bjelonic2022offline}.

%\red{However, extending this approach to locomotion systems introduces significant challenges. Their inherent complexity—arising from intricate dynamics and actuation configurations—makes real-time motion planning impractical. For example, forward or rotational motions of three-link swimmers, illustrated in \cref{fig:GraphicalAbstract}(a-b), result from complex interactions between joint oscillations and their environment. Hierarchical maneuver-based planning offers a solution to mitigate the computational burden in real-time scenarios \cite{frazzoli2001robust,frazz2005maneuverbased,niu2013locomotion,da20162d,bjelonic2022offline}. This approach separates planning into low-level and high-level components: the low-level motion planner generates a set of maneuvers (described by gaits) offline, based on an accurate model, to avoid real-time constraints. Maneuvers allow the desired body trajectories to be easily translated into corresponding joint trajectories. High-level planning, performed online, generates high-level trajectories (e.g., center of mass trajectories) by selecting and blending sequences of maneuvers.}

For effective high-level motion planning, the offline-constructed gait library must contain sufficient gaits to enable the system to reach any desired position (i.e., ensure controllability \cite{murray1993nonholonomic}) while minimizing the number of sequences needed (i.e., maximize maneuverability \cite{hatton2011geometrica,hatton2013snakes}). By extension, a continuous set of gaits—termed a \emph{gait family}~\cite{deng2022enhancing,choi2022optimal}—can further enhance both controllability and maneuverability.

To the best of our knowledge, most existing works on maneuver-based planning have focused on implementing high-level motion planning using gait libraries \cite{nguyen2020dynamic,da20162d,erez2011infinitehorizon} or generating suboptimal gait families via central pattern generators \cite{niu2013locomotion,ijspeert2008central}.  In this paper, we propose a framework to generate families of optimal gaits for steering kinematic locomoting systems without individually optimizing each gait, and to analyze the system's controllability and maneuverability as defined by optimal gait families. This framework and its application to a representative system are outlined in \cref{fig:GraphicalAbstract}:
\begin{enumerate}
\item A mobile system's maneuverability (analogous to the manipulability of a fixed-base arm) is characterized by the set of velocities the system can achieve at a given level of effort~\cite{hatton2011geometrica}. For example, as illustrated in \cref{fig:GraphicalAbstract}a, if we take the effort of moving a differential-drive car as the squared sum of its wheel speeds, the unit-effort body velocities trace out a constant-radius arc connecting pure-forward motion to pure-rotational motion. Similarly, we can generate an equivalent curve for a family of gaits by dividing their per-cycle displacement by the times required to complete the cycles at a given average power and note that the pure-turning and maximum-turning-rate points on this curve are not necessarily the same.
\item The displacement induced by a gait is determined by the signed-area it encloses on a function determined by the local system dynamics. As illustrated in \cref{fig:GraphicalAbstract}b, for a three-link swimmer, a pure-forward gait encloses a sign-definite region of the ``forward function" at the center of the joint-angle space, and a pure-turning gait encloses equal positive and negative regions of this function for a net-zero enclosure (while also enclosing a sign-definite region on a ``turning function" as discussed in~\S\ref{sec:background}). The time taken to execute a gait at normalized effort is determined by the weighted pathlength and curvature of the trajectory.
\item Varying the placement and shape of the gait curves in the joint-angle space produces a set of gaits characterized by the direction and speed of their average induced velocities. Projecting this set into the average-velocity space, as in \cref{fig:GraphicalAbstract}c, produces a set of points whose outer edge is the Pareto front containing the fastest gaits for each direction of motion, and thus the dashed curve from \cref{fig:GraphicalAbstract}a. Because the geometry of the gait dynamics in \cref{fig:GraphicalAbstract}b can produce distinct sets of gaits with the same average velocities, the projection has overlaps, which can result in difficulties such as bifurcation when using non-exhaustive optimization schemes to generate the Pareto front.
\end{enumerate}

The core of this paper explores the generation of the continuous steering Pareto front, of which the notion is introduced in~\cite{deng2022enhancing}, and the mitigation of the overlap and bifurcation complications. In particular, we employ a local search approach inspired by recent works on optimal gait families of bipedal robots in 1D settings~\cite{rosa2022topological,raff2022generating}, in which we first generate a ``seed gait" for the Pareto front, e.g., the optimal pure-forward gait, and then find nearby ``steering" gaits that mix forward and turning motion at different rates while maximizing average speed. This local search exploits nonlinear parametric programming methods~\cite{allgower2003introduction,jongen1990parametric,li2020optimization}, with continuations used to encode the changing gait-optimization criteria across the different steering rates. This approach allows us to generate a continuous family of steering gaits rather than simply optimizing gaits for a set of predetermined steering ratios. 

Additionally, we propose a global search approach as compensation for the local search approach. The local search approach needs information for the next step near the bifurcation points. Then, we use a brute-force search to find the gait satisfying the optimality condition in the reduced-order gait parameter space. This approach provides information about where the overlaps happen and the direction for local search near the bifurcation point.

Finally, a generalization of our search method also allows us to generate two-dimensional families of gaits, characterized by both steering rate and per-cycle step size. Including gaits of different step sizes in the family supports operations such as stationkeeping and deadbeat control: Optimal gaits at each steering rate have an associated ``stride length" that may not be compatible with the specified task. Having the option to take shorter ``baby steps" with which to respond to small disturbances, or longer ``lunges" to reach a position that is not an integral number of optimal cycles away from the starting position thus increases the accuracy with which the system can follow a reference command.

This paper is an extended version of \db{the works by Choi et al. \cite{choi2022optimal} and Deng et al. \cite{deng2022enhancing}.}{our previous works on gait families in \cite{deng2022enhancing,choi2022optimal}.}\db{Deng et al. \cite{deng2022enhancing}}{In one of these preliminary works \cite{deng2022enhancing}, we}explored maximum-efficiency steering gaits in the sense of Pareto optimality by nudging a forward gait into a turning gait or vice versa. However, this approach only generated a partial Pareto front in the average-velocity space because of a limitation of the weight-sum method \cite{kim2005adaptive}. \db{Choi et al. \cite{choi2022optimal}}{In the other work \cite{choi2022optimal}, we}identified the mathematical structure underlying an optimal gait family and built an Optimal Locus Generator to solve a family of optimization problems. Even though \db{Choi et al. \cite{choi2022optimal}}{we}verified that the Optimal Locus Generator successfully generated a family of ``baby-steps” producing smaller per-cycle displacements than the optimal gait (and thus allowing for finer granularity of motion), it could only handle a single constraint value and did not have a mechanism for handling the bifurcations that appear when optimizing steering gait families. In this paper, we explore the whole optimal gait family including baby-step gaits and gait with varying steering rates by searching for them locally and globally, based on the preliminary works.

\begin{figure*}
	\centering
	\includegraphics[width=\textwidth]{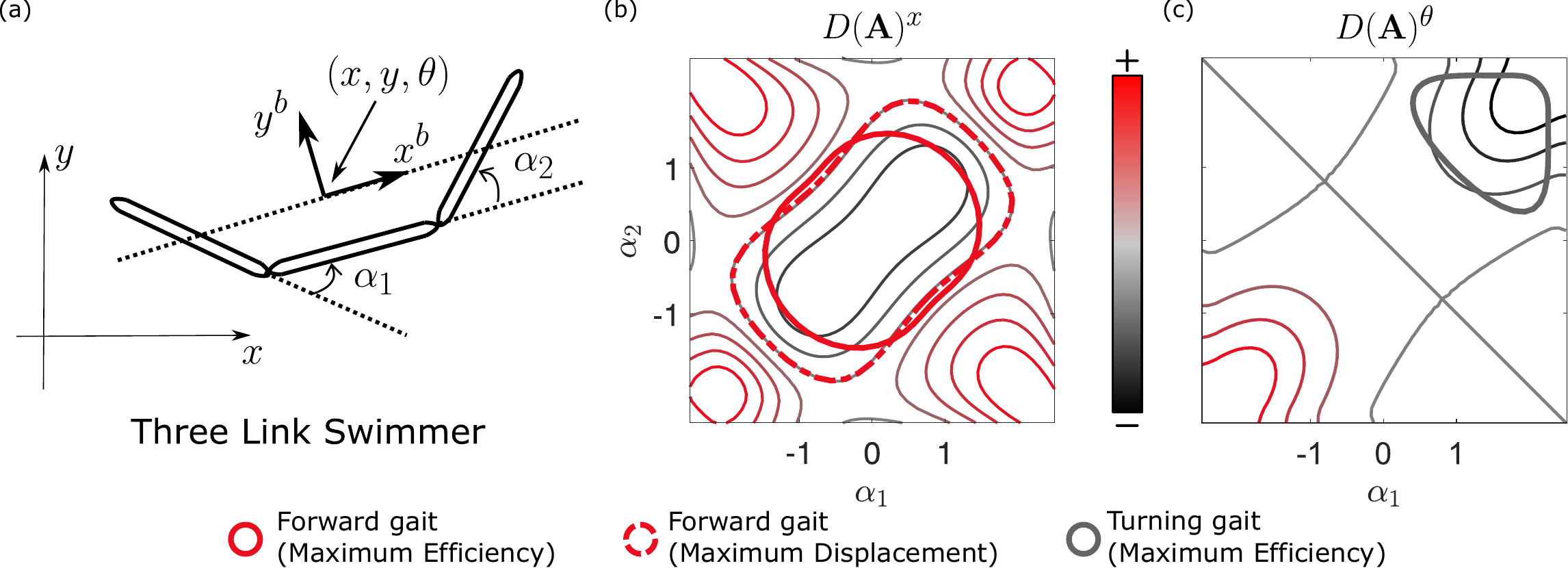}
	\caption{An overview of an example articulated swimmer and its optimal gaits. (a) Schematic of a three-link viscous swimmer. (b) Two optimal gaits: The maximum-displacement gait (red dashed line) in the $x$-direction, and the maximum-efficiency gait (red solid line) in the $x$-direction (i.e. the largest displacement in $x$-direction per unit power dissipation). The gait is plotted over the system’s \emph{Constraint Curvature} Function (CCF) corresponding to $x$-direction movement $D(\mixedconn)^x$, which gives an approximation of the system’s locomotive behavior. These two gaits are found in \cite{tam2007optimal,ramasamy2019geometry}. (c) The maximum-efficiency gait (gray solid line) for turning motion on the turning CCF.}
 % The shape of the system, described by its two joint positions, lives in a two-dimensional space. The coordinate of the body frame is chosen according to the \emph{minimal perturbation} principle.
	\label{fig:SystemExample}
\end{figure*}

To demonstrate our approach, we generate families of optimal gaits for steering three-link swimmers immersed in viscous \cite{purcell1977life} and perfect (inertial) fluids \cite{kanso2005locomotion,melli2006motion}. The three-link swimmer is a standard reference system in locomotion, and its dynamics have two properties that make the swimmer particularly useful for illustrating our approach: As illustrated in \cref{fig:GraphicalAbstract}b, the net displacement induced by a gait corresponds to the amount of constraint curvature that the gait encloses. For the viscous three-link swimmer, the time-energy cost of executing a gait corresponds to its metric-weighted pathlength through the shape space. For the perfect-fluid three-link swimmer, the actuator torque cost of executing a gait is related to the non-constraint (i.e.,\ covariant) acceleration as calculated from the generalized mass metric on the shape space, and on a second cometric that describes how the actuators are attached to the system \cite{cabrera2024optimal}.
%%%%%%%%%%%%%%%%%%%%%%%%% Background %%%%%%%%%%%%%%%%%%%%%%%%%

\section{Background}
\label{sec:background}
The analysis of locomotion in this paper builds on \db{the}{our and others'}previous works in geometric mechanics \cite{bloch1996nonholonomic, ostrowski1998geometric, kanso2005locomotion, avron2008geometric, shammas2007geometric,hatton2011geometrica, hatton2013geometric,ramasamy2019geometry,hatton2022geometry}. This framework provides both a rigorous formulation for and an intuitive description of how much the systems move given a particular gait cycle. The constraint curvature of dynamics in each direction, as illustrated in \cref{fig:SystemExample}b and \ref{fig:SystemExample}c, provides the locomotion information map on the shape space: the net displacement is approximated by the area-flux of a constraint curvature function (CCF) enclosed by the gait cycle. In the sense of the cost (i.e.,\ how much effort is needed to execute the gait), the dissipated energy and the required actuator force necessary to enact the gait were considered \cite{hatton2017kinematic, hatton2022geometry}.

\subsection{Representation of Position Space}
The configuration space $\configspace$ of a locomoting system can be separated into a shape space $R$ and a position space $\fiberspace$. For the planar articulated systems we consider in this paper, the systems’ shape spaces can be parameterized by their joint angles, e.g.,
$r=[\alpha_1,\alpha_2]^T$ for a three-link swimmer with two joints. The position $\fiberspace$ is the set of planar translations and rotations (the ``special Euclidean group'' in the plane), whose elements can be described in coordinates or matrix form as
\begin{equation} \label{eq:posrep}
    \fiber = (\fiberx, \fibery, \fibertheta) \qquad \text{or} \qquad \fiber = \SEmatrix{\fiberx}{\fibery}{\fibertheta}.
\end{equation}

The velocity of the system through the shape space is the vector of the rates at which the joint angles are changing, $\dot{r}=[\dot{\alpha}_1,\dot{\alpha}_2]^T$. The velocity of the system in the position space can be expressed as the rate at which the coordinates or corresponding matrix elements are changing,
\begin{equation} \label{eq:velrep}
    \fiberdot = \begin{bmatrix}\fiberdotx\\ \fiberdoty\\ \fiberdottheta\end{bmatrix} \text{   or   } \fiberdot = \SEmatrixvv{\fiberdotx}{\fiberdoty}{\fiberdottheta}{\fibertheta},
\end{equation}
(where the matrix components in~\eqref{eq:velrep} are tangent to the set of matrices in~\eqref{eq:posrep}, and the map from the column form of the vector to the matrix form is the \emph{representation} or \emph{hat} operation).

It is often more useful to work in terms of the system's body velocity $\fibercirc$ ($\fiberdot$ calculated in a frame for which the current configuration is at the origin), 
\begin{equation} \label{eq:bodyvelrep}
    \fibercirc = \begin{bmatrix}\fibercircx\\ \fibercircy\\ \fibercirctheta\end{bmatrix} \qquad \text{or} \qquad \fibercirc = \SEmatrixvvg{\fibercircx}{\fibercircy}{\fibercirctheta},
\end{equation}
(where we have replaced the time-derivative dot with an open circle to indicate that we are working in a local frame).\footnote{This is formally a Lie algebra element.} In particular, dynamic equations in homogeneous isotropic environments tend to be symmetric with respect to the system position and orientation, such that the position $\fiber$ can be factored out of the dynamics by working in local coordinates.

To integrate the body velocity into accrued displacement, we map it from the local frame to the global coordinate or matrix frame by the (lifted) action of the current position, $\fiberdot = \fiber\fibercirc$, which is expressed in coordinate form as
\begin{equation}
    \begin{bmatrix}\fiberdotx\\ \fiberdoty\\ \fiberdottheta\end{bmatrix} = \SEmatrixlLact{\fiberx}{\fibery}{\fibertheta}\begin{bmatrix}\fibercircx\\ \fibercircy\\ \fibercirctheta\end{bmatrix}\\
\end{equation}
(which rotates the position components to account for the current orientation of the system), or in matrix form as
\begin{equation}
\begin{split}
        &\SEmatrixvv{\fiberdotx}{\fiberdoty}{\fiberdottheta}{\fibertheta} = \\ & \qquad \SEmatrix{\fiberx}{\fibery}{\fibertheta} \SEmatrixvvg{\fibercircx}{\fibercircy}{\fibercirctheta} 
\end{split}
\end{equation}
(which rotates the position components and also rotates the orientation components to match the constraints on the matrix entries in~\eqref{eq:velrep}).

%The geometric locomoting systems have a configuration space $\configspace$, separated into a position space $\fiberspace$ and a shape space $R$. The elements in the shape space are parametrized by their joint angles, $r=\transpose{[\alpha_1,\alpha_2]} \in R$ for the three-link swimmer. In this paper, the position space $\fiberspace$ at the identity is the special Euclidean group of two-dimensional translations and rotations, and thus $\fiberspace = SE(2)$. The velocity space $\fiberlie$ is defined by the tangent space of the position space $\fiberspace$ at the identity, and thus $\fiberlie = \mathfrak{se}(2)$ (also known as the Lie algebra).  

% The action is defined with respect to the system's instantaneous body frame rather than the world frame. Thus, to describe the motion of the locomoting system, it is convenient to use the body frame. The body velocity $\fibercirc$ is the world velocity $\dot{\fiber}$ expressed in the instantaneous body frame:
% \begin{equation}
% 	\fibercirc = \fiber^{-1}\dot{\fiber}.
% \end{equation}

\subsection{Linear-kinematic locomoting systems}

For locomoting systems in drag-dominated, isolated-inertial, and perfect-fluid environments, the system equations of motion reduce to a set of constraints between their shape velocities and body velocities~\cite{shapere1989geometry, ostrowski1998geometric, ramasamy2019geometry}. If the shape changes are specified as control inputs, these constraints take the form of a linear map between shape and position velocities,
\begin{equation} \label{eq:localconnection}
	\fibercirc = \mixedconn(r)\dot{r},
\end{equation}
where $\mixedconn$ is variously referred to as a \emph{local connection}, \emph{locomotion Jacobian}, or \emph{motility map}, and is the same structure as appears when relating the motion of a simple differential-drive car to the rate at which the wheels are turning.\footnote{For historical reasons, many sources include a negative sign in the right-hand side of~\eqref{eq:localconnection}. We have removed it here to increase the clarity of the constraint curvature expression below.} In the coordinate representation, $\mixedconn$ takes the form of a standard matrix; in the matrix representation, each ``column" of $\mixedconn$ is itself a matrix with the same structure as in~\eqref{eq:bodyvelrep}.

The linearity of~\eqref{eq:localconnection} means that the position motion induced by a shape trajectory $\psi$ is kinematic (independent of time rescaling), such that the net induced motion $\fiber_{\psi}$ can be rewritten as a path integral in the shape space,
\begin{equation}
    \fiber_{\psi} = \int_{0}^{T} \fiber \mixedconn(r(t)) \dot{r}(t) \, dt = \int_{\psi} \fiber\mixedconn(r) \, dr,
\end{equation}
again producing either a set of coordinate variables or the corresponding matrix representation.

% body velocity is determined  their shape velocity $\dot{r}$ by the local connection $\mixedconn(r)$ that describes the interaction between the system and the environment (The method to derive the local connection can be found in \cite{hatton2013geometric, ramasamy2019geometry}):
% \begin{equation}
% 	\fibercirc = -\mixedconn(r)\dot{r}
% \end{equation}

\subsection{Gaits}

A gait is a cyclic trajectory $\gait$ in the shape of a system, e.g., an oscillation of its joint angles. The net displacement induced over the cycle depends on how $\mixedconn(r)$ changes across the cycle -- if the coupling is ``constant" over the shape space, then any motion achieved by moving ``away" from the initial shape is reversed when moving ``back" to that shape. 

The changes in the local connection that affect the net displacement are measured by the \emph{curvature} operator $D$, which for each pair of shape variables $i<j$ evaluates as
\begin{equation}
    D(\mixedconn)_{ij} = \left(\frac{\partial\mixedconn_{j}}{\partial r_{i}} - \frac{\partial\mixedconn_{i}}{\partial r_{j}}\right) + \liebracket{\mixedconn_{i}}{\mixedconn_{j}}.
\end{equation}
To distinguish $D(\mixedconn)$ from other kinds of curvature that appear in geometric analysis, we refer to it as the \emph{constraint curvature} function (CCF), highlighting that it ultimately measures how the system constraints vary across the configuration space.\footnote{Note that flipping the sign of $\mixedconn$ flips the sign of the first term but not the second term, so that $D(-\mixedconn)$ is not equal to either $D(\mixedconn)$ or $-D(\mixedconn)$. As mentioned above, we have removed the negative sign from~\eqref{eq:localconnection} to avoid a profiligation of negative signs in our exposition.}

The first (``nonconservative") term in the constraint curvature (the \emph{exterior derivative} which generalizes the notion of curl) measures changes in each component of $\mixedconn$ along \emph{other directions} in the shape space. Such changes mean that, for example one shape variable can be used to modulate the coupling between the other shape variable and position motion, allowing for ``large steps forward and small steps backward'' over a cycle.

The second (``noncommutative") term in the curvature (the \emph{Lie bracket}) measures to a first order how rotations induced through $\mixedconn$ by changing one shape variable ``redirect'' the global expression of the position motions induced by changing the other shape variable. These redirections allow a system to access actions that produce net displacement from motions that are locally reciprocal, e.g., the classic ``parallel-parking" action of moving forward and backward while turning left and right out of phase from the forward-backward motion. In generalized coordinates, the Lie bracket evaluates as
\begin{equation}
    \liebracket{\mixedconn_{i}}{\mixedconn_{j}} = \begin{bmatrix}
        \mixedconn^{y}_{i} \mixedconn^{\theta}_{j} - \mixedconn^{y}_{j} \mixedconn^{\theta}_{i}\\
        \mixedconn^{x}_{j} \mixedconn^{\theta}_{i} - \mixedconn^{x}_{i} \mixedconn^{\theta}_{j}\\
        0
    \end{bmatrix}
\end{equation}
and in the matrix representation it is equal to the difference between the matrix products of the ``columns" of $\mixedconn$ taken in different orders,
\begin{equation}
    \liebracket{\mixedconn_{i}}{\mixedconn_{j}} = \mixedconn_{i}\mixedconn_{j} - \mixedconn_{j}\mixedconn_{i}.
\end{equation}

By a generalization of Stokes' theorem, the net displacement induced by a gait cycle can be approximated by integrating the constraint curvature over a surface in the shape space bounded by the gait. As discussed below, this property means that we can look for ``good" gaits in regions of the shape space with strongly sign-definite curvature; it also means that optimal gaits lie along curves where the extra curvature that could be enclosed by varying the gait is in equilibrium with the extra cost that the variation would incur.\footnote{This second point is important, because it means that we do not in general have to identify a surface bounded by the gait, and instead use the fact that it could in theory exist.}

Specifically, the surface integral of the constraint curvature over a gait approximates the \emph{exponential coordinates} $\gaitexpdisp$ of the net displacement,
\begin{equation}		
	\gaitexpdisp \approx \iint_{\gait_{a}} D(\mixedconn), 
	\label{eq:BVI}
\end{equation}
where the exponential coordinates can be understood as body velocity which, if held constant for unit time, would bring the system from its initial configuration to its final configuration. Equivalently (and more usefully for our purposes), given a gait with cycle period $T$ and exponential coordinate displacement $\gaitexpdisp$, the quantity $\gaitexpdisp/T$ describes the average body velocity produced by the gait, and has the same coordinate or matrix structure used to describe $\fibercirc$ and $\mixedconn$.

The error in this approximation appears because the Lie bracket term in the constraint curvature only captures the nonlinear rotation-translation noncommutativity to first order~\cite{hatton2015nonconservativity,bass2022characterizing}. For any given system, it is possible to minimize this error by using system coordinates that project the nonlinearity of the constraints into the nonconservative portion of the curvature~\cite{hatton2011geometrica,hatton2015nonconservativity}. For locomoting systems, this choice of coordinates is a generalization of placing the body frame of a free-floating system at its center of mass and mean orientation.

\subsection{Cost of moving}

For our systems (shape-controlled viscous-swimmers, perfect-fluid swimmers, and free-fall systems), the same operations that allow us to reduce the equations of motion to a map from shape velocity to body velocity also allow us to equip the system shape spaces with Riemannian metrics $\mtensor$, such that the metric norm of the velocity
\begin{equation}
    \norm{\dot{r}}_{\mtensor}^2 = \transpose{\dot{r}} \mtensor \dot{r} = \innerprod{\dot{r}}{\dot{r}}_{\mtensor}
\end{equation}
on a viscous system, or of the covariant (or active) acceleration on a perfect-fluid or free-fall system,
\begin{equation}
    \norm{\covacc}_{\mtensor}^2 = \innerprod{\covacc}{\covacc}_{\mtensor}, \qquad \covacc = \ddot{r}-\inv{\mtensor}\coriolis_r(r,\dot{r})
\end{equation}
describes the cost of moving through the shape space~\cite{shapere1989efficiencies, avron2008geometric, hatton2017kinematic, ramasamy2019geometry, hatton2022geometry} in terms of the overall drag on or active acceleration of the system's constituent components $\covacc$, where $\coriolis_r(r,\dot{r})$ corresponds to the centrifugal and Coriolis forces acting on the system.

As discussed in~\cite{hatton2022geometry, cabrera2024optimal}, these metrics can be augmented with information about the leverage the actuators have on the system components and each other. The details of this formulation involve some differential geometric subtlety, but a key result is that squaring the metric tensor for the norm calculations (but not the covariant-acceleration calculation) measures the cost of moving or accelerating through the shape space in terms of the squared actuator force required for the motion.

The metrics we use are specifically constructed by first pulling back local metrics $\mu$ (e.g., fluid drag values or mass matrices) defined on system components through the system kinematics,
\begin{align}
    \left\lVert\begin{bmatrix}\fibercirc \\ \dot{r}\end{bmatrix}\right\rVert_{M_{\text{full}}}^2 &= 
    \innerprod{\begin{bmatrix}\fibercirc \\ \dot{r}\end{bmatrix}}{\begin{bmatrix}\fibercirc \\ \dot{r}\end{bmatrix}}_{M_{\text{full}}}
    \\
    M_{\text{full}} &= \sum_{i} J_{i}^{T} \mu_{i} J_{i}
% \begin{bmatrix}\transpose{\fibercirc} & \transpose{\dot{r}}\end{bmatrix} M_{\text{full}} \begin{bmatrix}\fibercirc \\ \dot{r}\end{bmatrix} 
    \label{eq:metriccalc}
\end{align}
where $i$ indexes the system components, and the $J$s are the Jacobians of those components with respect to the general system motion. The local connection can then be used to ``fold down" this full metric into a metric on the shape space that accounts for the induced position motion,
\begin{align}
    \norm{\dot{r}}_{\mtensor}^{2} &=  \innerprod{\dot{r}}{\dot{r}}_{\mtensor}
    \\
    \mtensor &= \begin{bmatrix}\transpose{\mixedconn} & \transpose{\matrixid}\end{bmatrix} M_{\text{full}} \begin{bmatrix}\mixedconn \\ \matrixid \end{bmatrix}.
\end{align}

These calculations are easiest to make when working with the coordinate representation of $\fibercirc$ (for which $\mtensor$ will be a standard matrix), but it is possible to construct an equivalent structure when working with a matrix-represented $\fibercirc$, e.g. by incorporating a ``flattening" operation that unpacks the matrix representation from $m\times m$ to $m^2 \times 1$ and compensates for any double-counting in the rotation terms.

Details of the integrals we take of these metrics are provided in \cref{app:gaitcost}.

%When evaluating system gaits, it is important to consider not only how far the gait moves the system, but how much the gait costs in terms of opportunity (e.g., how long does it take to execute the gait?) and effort (e.g., how much energy must the system expend on executing the gait?). For many systems (including the classes of systems we consider here), opportunity and effort costs are interchangeable -- maximizing speed for a given average effor

%The cost of executing a trajectory 

%Common costs include some notion of opportunity (e.g., how long does it take to execute the gait?) and effort (e.g., how much energy must the system expend on executing the gait?). To normalize the effort in the shape space, we can stretch or weight the shape space by some metric tensors. For drag-dominated system, the metric-weighted pathlength through the shape space, $\gaitcost$, is associated with the power cost of executing a gait. For inertia-dominated systems, the metric-weighted pathlength is associated with the kinetic energy the system must possess to complete a gait in a given time. To consider the second-order effect of dynamics, we can substitute acceleration- or torque-based cost metrics that additionally account for speed changes, path curvature, and the leverage the actuators have on the system masses. The detail of the gait cost used in this paper is discussed in and \cite{hatton2022geometry}.

\section{Gait Optimization}
\label{sec:gaitopt}
In previous works \cite{tam2007optimal, ramasamy2019geometry, hatton2022geometry}, \db{the geometric mechanic community}{we and others}have focused on optimizing gaits to maximize efficiency, both with and without constraints on specific motion patterns. These studies have presented gait optimization methods to systematically identify such gaits. For simple motion plans, such as straight-line trajectories or constant-curvature arcs, repeating a single optimal gait suffices. However, more complex tasks—such as obstacle avoidance or navigating highly curved paths—require a diverse set of gaits with varying step sizes and steering rates. This broader set of useful motions can be obtained by solving a family of optimization problems with different constraints on step size and steering rate.

\begin{figure*}[ht]
	\centering
	\includegraphics[width=\textwidth]{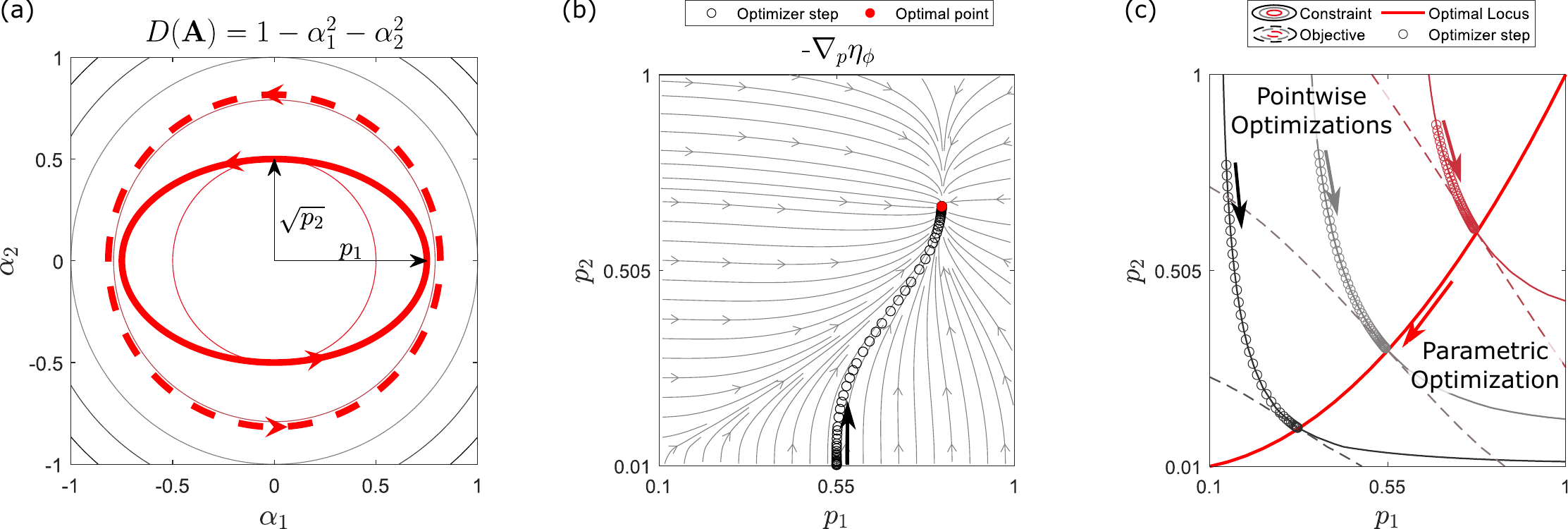}
	\caption{Example System for Gait and Gait Family Optimization (a) A level set of the CCF is depicted as a circular paraboloid, \(1 - \alpha_1^2 - \alpha_2^2\). For simplicity, the gait cycle is restricted to an ellipse, parameterized by \(p_1\) (the semi-major axis) and \(p_2\) (the square of the semi-minor axis). The unconstrained optimal gait is indicated by the red dotted line. (b) Schematic of unconstrained gait optimization. Each point in the parameter space corresponds to an elliptical gait cycle in the shape space. In the absence of constraints, the optimization proceeds by flowing along the vector field (e.g., the negative gradient of gait efficiency). The black circle represents an optimizer step, and the filled red circle marks the optimal point. (c) Schematic of constrained gait optimization. Here, the optimal points form a continuous curve where a level set of the cost function osculates with a level set of the constraint (net displacement) at the optimal point. Arrows indicate the search direction for solutions. Two solution methods are illustrated: pointwise optimization (with the circle representing an optimization step) and parametric optimization (with the thick red line representing a continuous solution curve—i.e., an optimal gait family). Dashed and solid lines denote level sets of the cost and the constraint, respectively.}
	\label{fig:LagrangeExample}
\end{figure*}

A straightforward approach to this family of optimization problems is to perform point-wise optimizations, generating a discrete set of gaits known as a gait library. The circle-marked paths in \cref{fig:LagrangeExample}c illustrate this method. While this approach utilizes our established gait optimization techniques \cite{ramasamy2019geometry, hatton2022geometry}, it results in a discretized collection of gaits. As a more comprehensive alternative, we consider numerical continuation methods \cite{allgower2003introduction, raff2022generating, rosa2022topological}. By formulating the family of optimization problems as parametric optimization problems—where step size or steering rate serves as a parameter—this approach generates a smooth, continuous manifold of gaits spanning the desired parameter range. The red arrow in \cref{fig:LagrangeExample}c represents the progression of parametric optimization. We refer to this solution manifold as a gait family. In this section, we briefly review the geometric interpretation of gait optimization \cite{tam2007optimal, ramasamy2019geometry, hatton2022geometry, choi2022optimal} and propose an extension to parametric optimization.

\subsection{Pointwise Gait Optimization}
In this subsection, we assume for simplicity of notation (and without loss of generality) that displacement can be projected down to a single “interesting” direction. We parameterize the gait cycles in $\allgaits$ with a set of parameters $\gaitparam$. The definition of the gait parameter that we used in this paper is presented in \cref{app:gaitparam}. 

\subsubsection{Unconstrained Gait Optimization}

The most important gaits, especially for motion where the distance to be traversed is longer than the locomoting system, are those that move most efficiently. The structure of the gait optimization problems is qualitatively similar to a weighted area-perimeter problem in which the goal is to enclose as much rich area as possible in a field whose quality diminishes with radius while minimizing the perimeter of the encircling curve. This quality can be measured by the gait efficiency defined as the ratio of the net displacement $\fiber_{\gait}$ to the associated cost $\gaitcost$:
\begin{equation}
	\gaiteff = \gaitefffrac.
\end{equation}
Then, a gait optimization problem is simply formulated by maximizing the gait efficiency:\footnote{Note that we do not express the inequality constraint explicitly in most optimization problems in this paper. However, the shape boundary (i.e., the joint limit) always serves as the inequality constraint. The formal derivation of the shape boundary is presented in \cref{app:jointlimit}.}
\begin{equation}
	\unconstmax{\gaitparam}{\gaiteff}.
	\label{eq:maxeffgaitdef}
\end{equation}

The optimization problem is identified to find the gait satisfying the below optimality condition:
\begin{equation}
	\grad[\gaitparam] \gaiteff = \frac{1}{\gaitcost} \left( \dgaitefffrac{\gaitparam} \right) = \mathbf{0},
	\label{eq:unconstrainedflow}
\end{equation}
i.e., an optimal gait is one for which any gains in displacement that can be achieved by varying the parameters are in equilibrium with the extra costs incurred. Note that the opposite direction of the gait efficiency's gradient $-\grad[\gaitparam]\gaiteff$ defines a vector field that can be flowed along to find a solution to~\eqref{eq:unconstrainedflow} if we assume that this optimization problem is a convex optimization problem. The solution is simply the point for which the gradient of gait efficiency is zero \cite{ramasamy2019geometry}.

\emph{Example: Weighted area-perimeter problem.}
For many systems, induced displacement increases with amplitude (bigger shape changes push the system further), but at a diminishing rate (leverage against the environment decreases at extreme angles, so the system gets less displacement per amount of shape change). By a similar token, incurred cost tends to increase monotonically with gait amplitude (large cycles take more time to execute at a given shape velocity, or take more shape velocity---and thus power---to execute in a given time).

For instance, suppose that the CCF is a ``hill'' with positive value at the center of the shape space and negative value where the radius is greater than 1, 
\begin{equation}
	D(\mixedconn) = (1-\alpha_1^2-\alpha_2^2),
 \label{eq:optexsys}
\end{equation}
and the associated cost is the pathlength (i.e.,\ the metric in this example is the Euclidean metric.) As illustrated in \cref{fig:LagrangeExample}a, the closed curve maximizing the ratio of enclosed quality to the perimeter is the dashed circle, which lies at the point where the diminishing quality of the field is balanced against the cost of expanding the circle to enclose more area. This problem as an unconstrained optimization can be solved by the gradient descent method which is stepping in the opposite direction of the objective gradient. As illustrated in \cref{fig:LagrangeExample}b, the gradient descent method can be viewed as a method to make flow from the initial guess into the vector field of the negative gradient.

\subsubsection{Constrained Gait Optimization}
Although maximum-efficiency gaits are important for moving a system over long distances, a gait-based planner may also need access to gaits for which the induced displacement is smaller than that of maximum-efficiency gaits satisfying \eqref{eq:maxeffgaitdef} to allow for following specified paths with more accuracy, or for station-keeping without shape drift. These gaits, which we call \emph{baby-step gaits}, are solutions to the optimization problem
\begin{equation}
	\constmaxone{\gaitparam}{\gaitefffrac}{\fiber_{\gait} = \contpar,}
	\label{eq:soptgaitdef}
\end{equation}
where $\contpar$ is a specified induced displacement.

To find sets of gait parameters satisfying this condition, we first note that because $\fiber_{\gait}$ is fixed,~\eqref{eq:soptgaitdef} reduces to the problem of finding the minimum-cost gait for a given displacement,
\begin{equation}
	\constminone{\gaitparam}{\gaitcost}{\fiber_{\gait} = \contpar,}
	\label{eq:soptgaitdef2}
\end{equation}
making this optimization specifically a weighted iso-areal problem. To find the optimum, we introduce $\lambda$ as a Lagrange multiplier, and use it to define a Lagrangian function $\lagrfun$ over the gait parameters,
\begin{equation}
	\lagrfun(\gaitparam,\lambda) = \alnth_{\gait} + \lambda(\fiber_{\gait}-\contpar).
	\label{eq:Lagrangian}
\end{equation}
For convenience, we assume that the gait is a local optimum if it satisfies the Karush-Kuhn-Tucker (KKT) conditions.\footnote{The KKT conditions are the first-order necessary optimality conditions, which is a generalized version of a Lagrange multiplier method. In general, they are not sufficient for optimality, thus the sufficient optimality condition is needed to check the gait optimality. The formal definition of the KKT conditions is described in \cref{app:optimization}.} The parameters of the baby-step gait can then be found by solving for the point $(\gaitparam^*,\lambda^*)$ where the derivative of $\lagrfun$ with respect to both the gait parameters and the Lagrange multiplier is zero,
\begin{equation}
	\nabla_{\gaitparam,\lambda} \lagrfun = \mathbf{0}.
	\label{eq:constrainedflow}
\end{equation}
Unlike the unconstrained optimization case, the solution to this problem cannot be found by flowing along the vector field defined by the left-hand side of \eqref{eq:constrainedflow}, because the solution $(\gaitparam^*,\lambda^*)$ (i.e.,\ the stationary point of the vector field) occurs at a saddle point of the Lagrangian function.

\emph{Example: Weighted Isoareal Problem.} The constrained version of the weighted area-perimeter problem is the weighted isoareal problem: find the shortest closed curve that encircles a given weighted area. We consider the same CCF and the cost metric in \eqref{eq:optexsys} as we used for a weighted area-perimeter problem. The contours in \cref{fig:LagrangeExample}c represent level sets of the cost function (dashed) and net displacement (solid). The points where these curves touch minimize the cost along the displacement level set and maximize the area per perimeter on that level set. The constrained optimizer forces the step to move on the feasible space, as shown in \cref{fig:LagrangeExample}c.\footnote{Although there exist various methods for constrained optimization (e.g., interior point method, active set methods, or sequential quadratic programming \cite{bazaraa2006nonlinear,boyd2004convex}), we give an example by using the projected gradient method here for a geometric intuition.}

%%%%%%%%%%%%%%%%%%%%%%%%% Optimal gait in one direction %%%%%%%%%%%%%%%%%%%%%%%%%
\subsection{Family of Gait Optimizations}
\label{subsec:famopt}

A family of optimization problems can be formulated as nonlinear parametric programming by introducing a parameter $\contpar$ into the optimization process \cite{jongen1990parametric}. While conventional optimization yields a single optimal point, parametric optimization defines a solution manifold parameterized by $\contpar$. To track solutions across this family, we reformulate the optimality conditions (including inequality constraints) as a system of nonlinear equations and apply numerical continuation methods to trace the solution manifold from an initial point.

Among various numerical continuation techniques, we adopt the predictor-corrector method, which effectively handles special points on the solution curve, such as bifurcations or jumps. In the predictor step, the method estimates the next point on the solution curve based on the current solution. The corrector step then refines this estimate using the Newton-Raphson method to ensure convergence to the true solution. In this work, we focus on our modified predictor step, which incorporates redundancy handling for gait parameters. A detailed discussion of the full predictor-corrector continuation method can be found in \cite{keller1978global, chan1982arclength, allgower2003introduction}.

\subsubsection{Problem Formulation}
In our analysis, the continuation function, $\contfcn(\contvar,\contpar)$, is the reformulation of the optimality conditions (specifically, KKT conditions) for the gait family. The continuation parameters, $\contpar$, change the structure of problems (e.g., the whole structure of the optimization problem in \eqref{eq:soptgaitdef2} is changed by handling the displacement level $\contpar$ as a continuation parameter.) Consider a system of $n$ nonlinear equations in $n+k$ variables:
\begin{subequations}
\label{eq:convnumcont}
\begin{equation}
	\contfcn(\contvar,\contpar) = 
	\begin{bmatrix}
		\grad[\gaitparam] \lagrfun \\
		\objmult^2 + \norm{\eqmult}^2 + \norm{\ineqmult}^2 - 1 \\
		\ncpfunset(\ineqmult,-\ineqfun) \\
		\eqfun(\gaitparam,\contpar) \\
	\end{bmatrix} = \mathbf{0},
	\label{eq:contkktdef}
\end{equation}
for
\begin{equation}
	\ncpfunset(\ineqmult,-\ineqfun) = \transpose{\begin{bmatrix}
		\ncpfun(\ineqmult[1],-\ineqfun[1]) & \cdots & \ncpfun(\ineqmult[\numineq],-\ineqfun[\numineq])
	\end{bmatrix}},
\end{equation}	
\end{subequations}
where $\contvar = \transpose{[\gaitparam,\objmult,\eqmult,\ineqmult]} \in \contvarspace$ are continuation variables including the gait parameter $\gaitparam$ and the dual variables (i.e.,\ Lagrange multipliers for objective function $\objmult$, equality constraints $\eqmult$, and inequality constraints $\ineqmult$), $\contpar \coloneqq \transpose{[\contpar_1, \cdots, \contpar_k]} \in \contparspace$ is continuation parameter, $\contpar_i \in (\cimin,\cimax)$, $\contfcn:\contvarspace \times \contparspace \rightarrow\euclid^n$, and $\ncpfun$ is a Fischer-Burmeister function.

The first row of the continuation function means the first-order optimality condition, the second row normalizes the dual variables, the third row describes the feasibility associated with the inequality constraint, and the fourth row describes the feasibility associated with the equality constraint. In particular, the inequality equations make it difficult to build a system of equations. To replace the inequality constraints with a system of nonlinear equations, we use the Fischer-Burmeister function $\ncpfun$ \cite{fischer1992special,fischer1997solution}. The formal definition of the Fischer-Burmeister function is described in \cref{app:numcont}.

The fundamental problem is how to track the fixed point of the nonlinear system (e.g., the optimal gait) as the continuation parameter  $\contpar$ (e.g., the step size or the steering rate of the gait) changes. The solution manifold contains the points satisfying \cref{eq:convnumcont}, $\contvar^*$, and the subspace only containing the gait parameter space represents the gait family. 

\subsubsection{Null-Projection Tangential Predictor}
In our problem setting, the optimal gait families form high-dimensional manifolds rather than curves because of the redundancy from the higher-order gait parameterization. For example, a Fourier parameterization of a gait cycle can generate gaits of the same shape but with different phases in the cycle, and these multiple gait parameters are interchangeable with each other. We do not consider this redundancy as a bifurcation.

\begin{figure}
	\includegraphics[width = 0.45\textwidth]{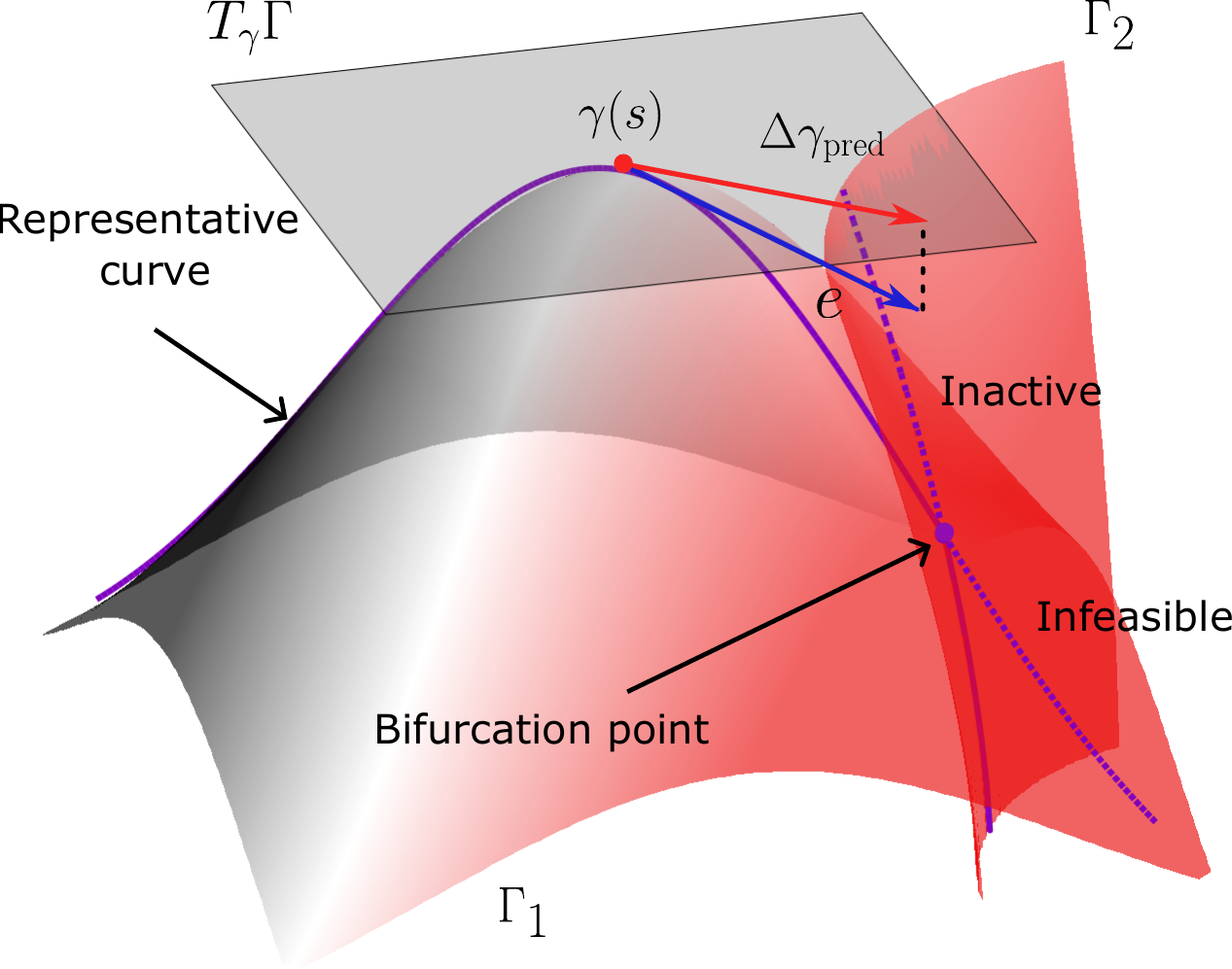}
	\centering
	\caption{The manifold $\contsolspace$ consists of two subsets, $\contsolspace_1$ and $\contsolspace_2$, each representing all possible gait parameters, though some correspond to qualitatively identical gaits. If $\contsolspace_1$ contains qualitatively distinct gaits from $\contsolspace_2$, a bifurcation occurs at their intersection. The dashed and solid curves represent different solutions from separate branches. The surface contour color represents the continuation parameter $\contpar$. $\contpredstep$ is the tangent predictor, $\contsearchvec$ is the unit vector in the search direction, and $\contsoltanspace[\contsol]$ denotes the tangent space of the solution manifold at $\contsol$.}
	\label{fig:PredStep}
\end{figure}

Building on the ideas above, our aim is to generate a curve through the space of gaits in which each point identifies a unique cycle. We specifically use the tangent predictor for estimating the next solutions. The solution manifold is the collection of points in a combination of a continuation variable and parameter space satisfying a system of equations $\contfcn$,
\begin{equation}
	\contsolspace \coloneqq \left\{\contsol \in \contvarspace \times \contparspace \ | \  \contfcn(\contsol) = 0 \right\},
\end{equation}
and $\contsol \subseteq \contsolspace$ be the solution curve parameterized by the arc-length. The tangent space of the solution manifold is the subset of the null space of the Jacobian of the continuation function \cite{hubbard2015vector}:
\begin{equation}
\contsoltanspace[\contsol] \subset \nullspace(\grad[\contsol] \contfcn) \coloneqq \nullcont,
	\label{eq:predstep1}
\end{equation}
where $\nullspace$ is a null space, $\contsoltanspace[\contsol]$ is the tangent space of the solution manifold $\contsolspace$ at $\contsol(\alnth)$ for $\alnth \in I$. 

To avoid the gait parameter redundancy, we propose a null-projection tangential predictor, $\contpredstep$. It is derived by projecting the basis vector of continuation variable space, $\contsearchvec$, onto the null space of the Jacobian of the continuation function $\nullcont$:
\begin{equation}
	\contpredstep \coloneqq \text{Proj}_{\nullcont}\left(\contsearchvec\right).
	\label{eq:predstep2}
\end{equation}
Away from the bifurcation point, the tangential vectors of the solution curve mostly are aligned to the direction to increase (or decrease) the continuation parameter $\contpar$. Then, the corrector finds the actual point on the solution manifold via the Newton method. 

However, bifurcations accompanied by qualitative changes in gaits can also occur, typically due to multiple solutions or the activation of inequality constraints. As illustrated in \cref{fig:PredStep}, such bifurcations cause the solution curve to branch, making the tangent space undefined at the bifurcation point. For instance, inequality constraints that enforce joint angle limits can trigger branching when their status changes (e.g., from inactive to active). Suppose $\contsolspace_1$ represents the solution branch where the inequality constraint is inactive, and $\contsolspace_2$ represents the branch where it is active. After the bifurcation, solutions in $\contsolspace_1$ become infeasible, while those in $\contsolspace_2$ are feasible but were suboptimal before the bifurcation. Connecting these solution curves results in a non-smooth transition. To handle such bifurcations, we project the point onto the feasible region, ensuring the selection of a valid solution branch. The algorithm of the proposed predictor-corrector method is described in \cref{app:numcont}.

\begin{figure}
	\centering
	\includegraphics[width=0.5\textwidth]{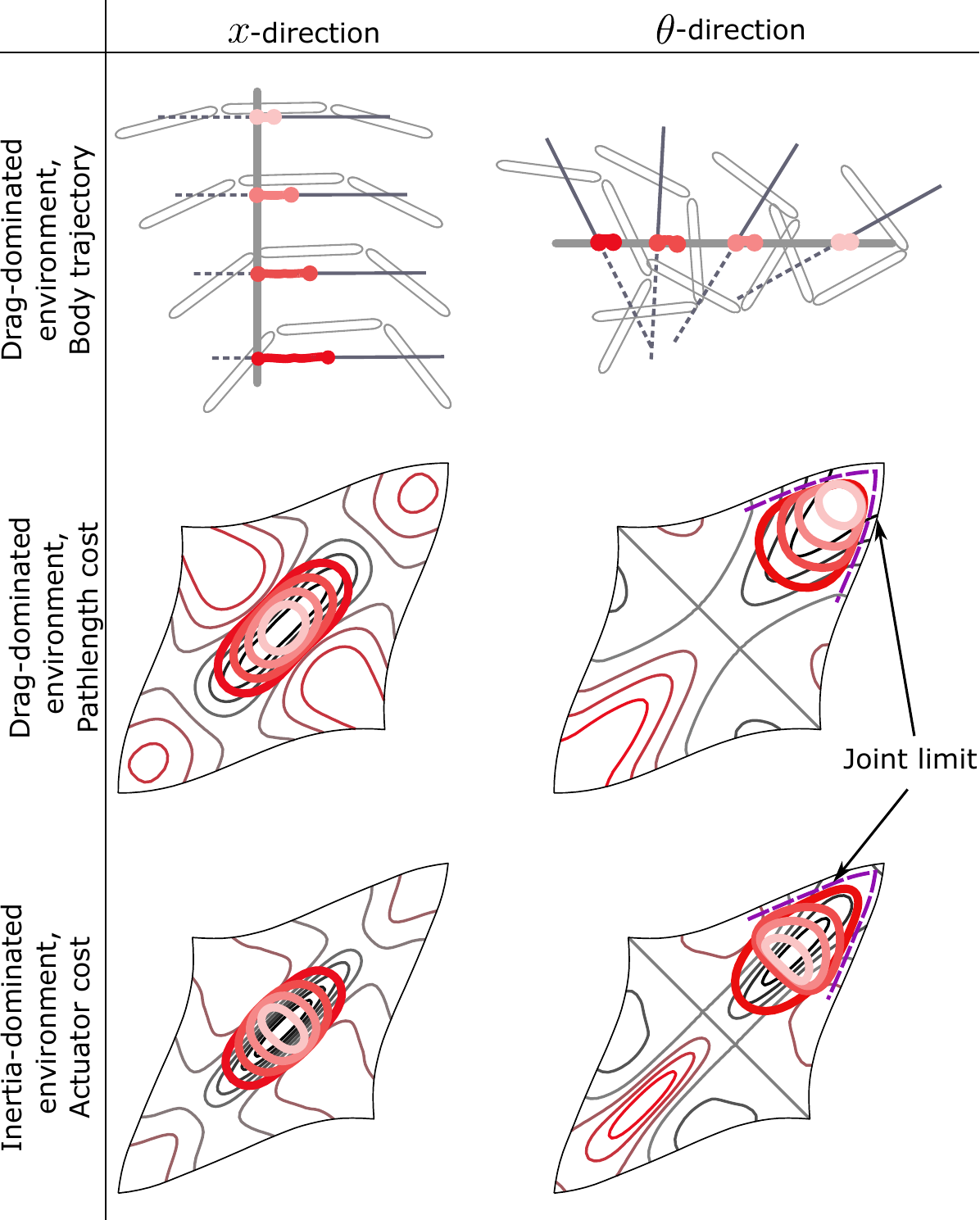}
	\caption{Translation and rotation gait families of the three-link swimmers under the drag-dominated and inertia-dominated environments. Our algorithm generates a continuous set of gaits producing different levels of displacement, of which we illustrate four each for translation and rotation, with $1, \frac{3}{4}, \frac{2}{4},$ and $\frac{1}{4}$ step size of the maximum efficiency gait. The first column shows the body trajectories of the viscous three-link swimmer when the system executes each gait three times. The second and third columns show the $x$ and $\theta$ baby-step gaits for the viscous and inertial three-link system in isometric embedding coordinates (for which metric pathlength is approximately the same as in-page pathlength), plotted against the corresponding $x$ and $\theta$ CCF contours. A detailed explanation of the isometric embedding space can be found in \cref{app:gaitcost} and \cite{hatton2017kinematic}.}
	\label{fig:StepOptGaitResult}
\end{figure}

\subsubsection{Baby-step Gait Family}
Using the proposed method, we generate baby-step optimal gaits for three-link swimmers immersed in both viscous and perfect-fluid environments. In this optimization problem \eqref{eq:soptgaitdef2}, the continuation parameter $\contpar$ represents the step size induced by the gait cycle.

The initial (seed) gait is chosen as the maximum-efficiency gait in each direction, corresponding to the net displacement $\contparmax$ (as illustrated in \cref{fig:SystemExample}b and \ref{fig:SystemExample}c).\footnote{The maximum-efficiency gaits for three-link swimmers in drag-dominated and inertia-dominated environments are detailed in \cite{tam2007optimal, ramasamy2019geometry, hatton2022geometry}.} The integration continues until the net displacement $\contpar$ reaches a specified minimum value $\contparmin$. This method yields an optimal gait family for motion in the $x$ and $\theta$ directions over a single cycle. Each gait within the family produces step sizes ranging from the maximum-efficiency displacement to near-zero motion. Four representative gaits are shown in \cref{fig:StepOptGaitResult}.

For the drag-dominated system, the power cost is associated with the gait’s path length, forming a constrained version of the weighted area-perimeter problem. The maximum-step gait encloses most of the rich area of the CCF while incurring some path-length cost. As the step size decreases, the gaits gradually become more circular, shifting toward the densest CCF region while sacrificing secondary-rich areas to reduce path length. Notably, turning gaits are constrained by joint limits since the highest-density region of the $\theta$-CCF is near these limits.

For the inertia-dominated system, the actuation-based cost is associated with the covariant acceleration of the shape trajectory. As a result, actuation-optimal gaits are more rounded than path-optimal gaits to minimize unnecessary acceleration. The actuation-based cost reflects the actuator's physical configuration, which in our case aligns with the joint configuration. Consequently, actuation-optimal gaits tend to be oval-shaped with a shorter extent along the $\alpha_1=\alpha_2$ axis, as motions along this axis (which curl the robot into a C-shape) cause the motors to backdrive each other, increasing cost.

%%%%%%%%%%%%%%%%%%%%%%%%% Steering optimal gait %%%%%%%%%%%%%%%%%%%%%%%%%
\section{Steering Gait Optimization}
\begin{figure*}
	\includegraphics[width=\textwidth]{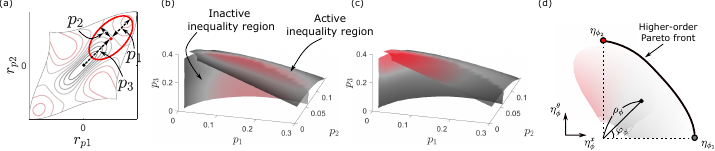}
	\caption{(a) Schematic of the elliptical gait parameterization. The ellipse is aligned along the $\alpha_1 = \alpha_2$ axis in isometric embedding coordinates ($r_{p_1}$, $r_{p_2}$), with semi-axes $p_1$ and $p_2$, and an offset $p_3$ from the origin to its center along that axis. (b–c) Gait family in the parameter space generated via the global search method. The surface is divided into regions where the inequality constraint is either active or inactive. In (b), the colormap represents Pareto efficiency—with red indicating higher efficiency and black lower efficiency. In (c), the colormap reflects the qualitative similarity to forward or turning gaits; black regions correspond to forward-like gaits, while red regions indicate turning-like gaits. (d) Gait family in the $x$–$\theta$ efficiency space from the global search method, overlaid with the Pareto front from the local search method. The polar coordinate parameterization facilitates the measurement of Pareto efficiency: the anchor points, corresponding to maximum forward and turning gaits ($\eff_{\gait_1}$ and $\eff_{\gait_2}$), serve as reference points. The radial coordinate $\paretorad$ represents Pareto efficiency (consistent with the colormap), and the angular coordinate $\paretoangle$ indicates the weighting factor describing the transition between the two anchor points.}
	\label{fig:GlobalApproach}
\end{figure*}

\label{sec:steering}

A common scenario in locomotion is steering during forward motion, where a robot adjusts its heading and speed to reach a target by executing a steering gait. Hierarchical maneuver-based steering control selects a gait parameterized by two action variables: forward velocity ($\forvel$) and rotational velocity ($\rotvel$). Together, these define gait efficiency as an $se(2)$ object:
\begin{equation}
	\gaiteff = \frac{\gaitexpdisp}{\gaitcost} =
	\begin{pmatrix}
		\forvel & 0 & \rotvel
	\end{pmatrix},
	\label{eq:steeringmotion}
\end{equation}
where gait efficiency is the ratio of the exponential coordinate of the net displacement ($\gaitexpdisp$) to the gait period (or cost) ($\gaitcost$), representing the average body velocity over a full gait cycle.

The goal of steering gait optimization is to minimize the gait period (or cost) while ensuring prescribed steering motions ($\forvel$ and $\rotvel$):
\begin{equation}
	\constminone{\gaitparam}{\gaitcost}{\gaitexpdisp =
	\begin{pmatrix}
		\contpar_1 & 0 & \contpar_2
	\end{pmatrix}.}
	\label{eq:steeringgaitdef}
\end{equation}
To solve this problem, we first conduct a global brute-force search to construct a reduced-order solution manifold, managing computational complexity. We then refine these solutions using nonlinear parametric optimization (a local search method), allowing for higher-order gait parameters and improved accuracy. However, numerical instability may arise near bifurcation points.

\subsection{Global Search Method}
To achieve a lower computational burden and simplify the analysis, we heuristically select reduced-order gait parameters. The reduction is guided by the following considerations:
\begin{itemize}
    \item Steering gaits should be a combination of forward and turning gaits.  
    \item The high-yield regions of both $x$-CCF and $\theta$-CCF are located along the even axis ($\alpha_1 = \alpha_2$).  
    \item In the isometric embedding space, an elliptical shape descriptor provides an approximate solution to a weighted isoareal problem, minimizing path length while maintaining a specified area.\footnote{We use a fourth-order Fourier descriptor for higher-order gait representation, with the elliptical shape descriptor serving as a reduced-order approximation. Further details are provided in \cref{app:gaitparam}.}  
\end{itemize}  

Based on these principles, we parameterize the gait using a tilted ellipse aligned with the even axis in the isometric embedding space. As illustrated in \cref{fig:GlobalApproach}a, this parameterization consists of three elements: the semi-axes of the ellipse, $p_1$ and $p_2$, and the offset $p_3$, which measures the displacement of the ellipse center along the even axis.  

We search for the optimal point satisfying the necessary conditions for \eqref{eq:steeringgaitdef}. Note that the stationarity condition for the optimization problem in \eqref{eq:steeringgaitdef} remains identical across optimal points, regardless of the continuation parameter
\begin{equation}
	\grad[\gaitparam]{\gaitcost}+\transpose{\eqmult}\grad[\gaitparam]{\gaitexpdisp} = 0
\end{equation}
To navigate the gait parameter space, we leverage the geometric interpretation of the stationarity conditions (or the Lagrange multiplier method): at an optimal point, the cost gradient can be expressed as a linear combination of the constraint gradients.

This approach eliminates the need to explicitly derive the dual variables. To see why, assume that the gradients of the constraints are linearly independent.\footnote{In this reduced-order problem, singularity does not occur because $x$-CCF and $\theta$-CCF are distinct.} Under this assumption, the optimality condition simplifies to:
\begin{subequations}
	\begin{equation}		
		\rank\left(
			\begin{bmatrix} 
			\grad[\gaitparam] \gaitcost & G
			\end{bmatrix}\right) = \rank\left(G\right),
	\end{equation}
	for
	\begin{equation}
		G = \begin{bmatrix}\grad[\gaitparam] \gaiteffx & \grad[\gaitparam] \gaitefftheta & \grad[\gaitparam] \ineqfun[\text{active}]
		\end{bmatrix},
	\end{equation}
	\label{eq:globalsteercond} 
\end{subequations}
where $\ineqfun[\text{active}]$ is the vector-value function representing each value of active inequality constraints. Then, $\grad[\gaitparam] \ineqfun[\text{active}]$ is the Jacobian transpose of $\ineqfun[\text{active}]$. If the gait does not reach the joint limit, $\ineqfun[\text{active}]$ remains empty in our problem setting.

We applied this method to generate a gait family for a viscous three-link swimmer.\footnote{This method is not suitable for inertia-dominated systems, as the reduced-order gait parameter lacks the ability to capture velocity or gait pacing—key characteristics for such systems.} \Cref{fig:GlobalApproach}b and \ref{fig:GlobalApproach}c illustrate the reduced-order optimal gait family in the gait parameter space, obtained by collecting points that satisfy the condition in \eqref{eq:globalsteercond}.

As illustrated in \Cref{fig:GlobalApproach}b, Pareto optimality is a function of the elliptical shape parameters ($\gaitparam_1$ and $\gaitparam_2$). Higher Pareto optimality results in greater motion per unit cost (e.g., maximum-efficiency gaits), while lower Pareto optimality produces less motion per unit cost (e.g., baby-step or lunge-step gaits). Meanwhile, \Cref{fig:GlobalApproach}c demonstrates that the offset parameter, $\gaitparam_3$, governs the ratio of forward to turning motion. As the offset increases, the gait generates more turning motion; as it decreases, the gait favors forward motion.

The surface in the parameter space is divided into two regions whether the inequality constraint is active or not, which is the bifurcation. As illustrated in \cref{fig:GlobalApproach}b, roughly speaking, the forward-motion-dominant-gaits are in the inactive inequality region, and the turning-motion-dominant-gaits are in the active inequality region. Bifurcation happens near the intersection of these two regions. Note that this result cannot fully represent the actual behavior of bifurcation in higher order solution manifold because we use the reduced order gait parameter.

\subsection{Local Search Method}
To derive a higher-order solution manifold, we apply the numerical continuation method we presented in \cref{subsec:famopt} to solve the steering optimization problems. Note that the steering gait family manifold is parameterized by the forward and rotational velocities, which means that the manifold is a parametric surface with two parameters topologically. However, this method generates a parametric curve with one parameter. Thus, we can divide the whole process to generate the solution surface into two steps of generating curves following each local coordinate of the solution surface. First, we generate a boundary of a solution surface (Pareto front). Then, we generate an interior of a solution surface (Baby gaits) starting from a point on the boundary.

\subsubsection{Pareto front of gait famlies}

The first step to generate the gait family in the higher-order parameter space is finding the boundary of the gait family in the $x$-$\theta$ efficiency space. The gaits on the boundary form the Pareto front, representing the best trade-offs between the motion efficiencies in the $x$ and $\theta$ directions, where we cannot improve one objective without hurting the other. Gaits on this front generate a range of rotational motion while largely preserving the overall translational motion. Thus, the structure of this problem is a multi-objective optimization problem to maximize efficiencies in the $x$ and $\theta$ direction:
\begin{equation}
	\unconstmax{\gaitparam}{\{\gaiteffx,\gaitefftheta\}},
	\label{eq:paretogaitdef}
\end{equation}
where $\gaiteffx$ and $\gaitefftheta$ are the gait efficiency in the $x$ and $\theta$ direction, generated by the gait cycle $\gait$, respectively. 

\begin{figure}
    \centering
    \includegraphics[width=0.8\columnwidth]{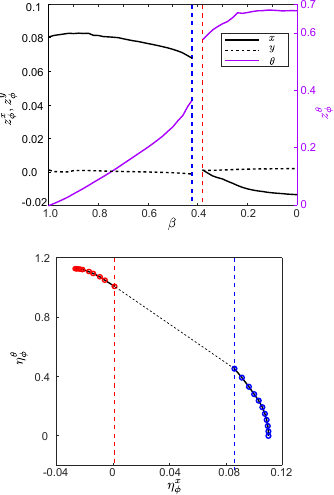}
    \caption{An overview of the performance of the weighted-sum method \cite{deng2022enhancing}. (a) the exponential coordinate of the net displacement per cycle of the steering gait on the Pareto front. Note that the discontinuity appears around the region $\beta=0.4$. (b) Pareto front, generated from \eqref{eq:wsumdef}, on the efficiency trade-off map. The discontinuity appears here as a large gap where the optimizer was not able to reliably converge to data points in between. The objective weight $\beta$ cannot distribute the Pareto optimal gaits evenly, and the Pareto optimal gaits are concentrated around each anchor point. }
	\label{fig:WeightedSum}
\end{figure}

\db{The preliminary study \cite{deng2022enhancing}}{In our preliminary study \cite{deng2022enhancing}, we }solved this problem in \eqref{eq:paretogaitdef} by the weighted-sum method. This approach generates a set of Pareto optimal gaits by optimizing the following objective while varying $\beta$ from 0 to 1:
\begin{equation}
	\gaiteff[\text{total}] = \beta \frac{\gaiteffx}{\anchoreff{1}^x} + \left(1-\beta\right) \frac{\gaitefftheta}{\anchoreff{2}^\theta},
	\label{eq:wsumdef}
\end{equation}
where $\gait_1$ and $\gait_2$ denote the optimal forward and turning gaits, respectively, and $\gaiteffx$ and $\gaitefftheta$ are normalized by their respective optimal values, with $\anchoreff{1}$ and $\anchoreff{2}$ serving as the anchor points. A drawback of this method is that the resulting Pareto front tends to concentrate around each anchor point (see \cref{fig:WeightedSum}), which is a well-known limitation of the weighted-sum method \cite{kim2005adaptive}.

In this work, we employ the numerical continuation method, reformulating the multi-objective problem as a family of single-objective problems. As illustrated in \cref{fig:GlobalApproach}d, Pareto optimization is conveniently expressed in polar coordinates. The origin is set at the intersection of vertical and horizontal lines passing through one of the anchor points. We normalize efficiency in the $x$- and $\theta$-directions by their respective maximum efficiencies and transform the point into polar coordinates. The radius $\paretorad$ represents the combined efficiency of forward-turning motion, while the angle $\paretoangle$ denotes the efficiency ratio between forward and turning motions.

Reformulating the Pareto optimization problem in \eqref{eq:paretogaitdef}, we seek to maximize the radius for specific angles, starting from the anchor point $\eff_{\gait_1}$:
\begin{subequations} 
\begin{equation} 
\constmaxone{\gaitparam}{\paretorad^2}{\paretoangle = \contpar,} 
\end{equation} 
where
\begin{gather} 
\paretorad^2 = \left(\gaiteffxnorm\right)^2 + \left(\gaiteffthetanorm\right)^2,\\ \paretoangle = \text{atan2}(\gaiteffthetanorm,\gaiteffxnorm), \\
\gaiteffxnorm = \frac{\gaiteffx-\anchoreff{2}^x}{\anchoreff{1}^x-\anchoreff{2}^x}, \
\gaiteffthetanorm = \frac{\gaitefftheta-\anchoreff{1}^\theta}{\anchoreff{2}^\theta-\anchoreff{1}^\theta}. 
\end{gather} \label{eq:steeringboundgaitdef} 
\end{subequations}
This formulation allows us to maximize the overall efficiency (as captured by \(\paretorad\)) for each prescribed efficiency ratio (\(\paretoangle\)), effectively capturing the trade-off between forward and turning motions.

\begin{figure}
    \centering
	\includegraphics[width=0.8\columnwidth]{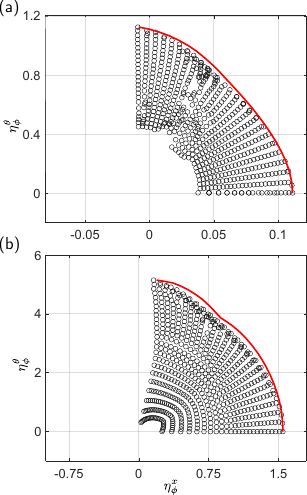}
	\caption{The steering gait family in the forward-turning efficiency space of (a) viscous three-link swimmers and (b) perfect-fluid three-link swimmers generated by the local approach. $x$- and $\theta$-axis respectively indicate the forward and turning efficiencies. The red curve indicates the Pareto front. The set of black circles indicates the gaits generating specific steering rates and step sizes. The gray dashed line over the figure indicates the vertical line corresponding to the pure rotation gaits. The surface indicates the optimal gait family generated by the global approach, and its colormap indicates the angle in the polar coordinate $\paretoangle$. Note that the global approach is not suitable for the inertia-dominated systems.}
	\label{fig:WholeGaitResults}
\end{figure}

We apply this method to determine Pareto fronts in efficiency space for three-link swimmers in both drag- and inertia-dominated fluids. The red curves in \cref{fig:WholeGaitResults}a and \ref{fig:WholeGaitResults}b depict the Pareto-optimal steering gaits for the viscous and perfect-fluid swimmers, respectively. For the viscous swimmer, forward gaits at higher-order parameters resemble the elliptical gaits obtained via the global search method. In contrast, elliptical turning gaits exhibit lower turning efficiency compared to higher-order turning gaits. This discrepancy arises because joint limits constrain turning gaits, preventing elliptical gaits from fully capturing the high-yield-$\theta$-CCF regions near the shape boundary.

\begin{figure}
	\centering
	\includegraphics[width=0.5\textwidth]{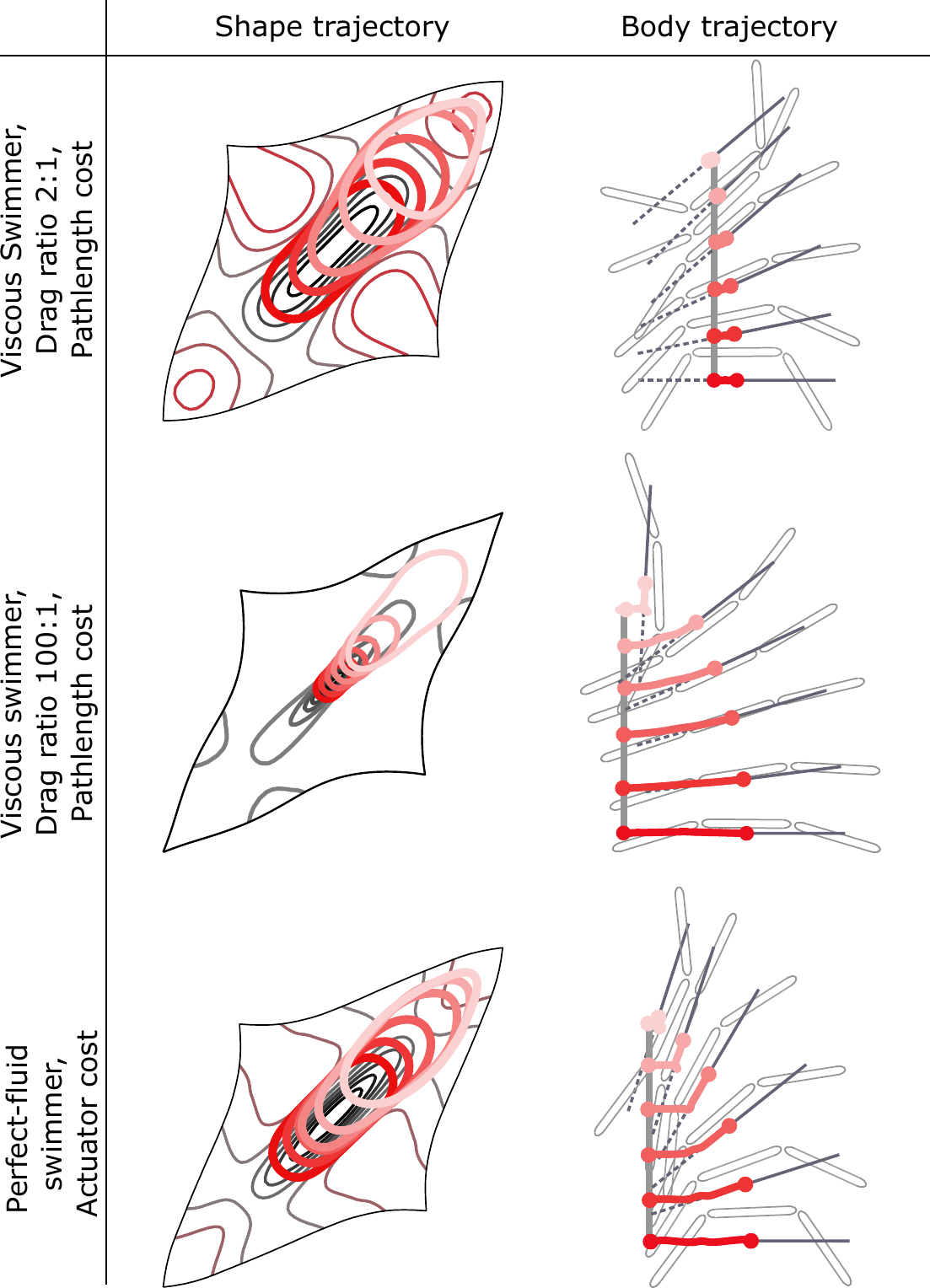}
	\caption{Pareto optimal steering gaits families of the three-link swimmers under the drag-dominated (first and second rows) and inertia-dominated (third row) environments. In particular, we explored the drag-dominated systems immersed in two different viscous fluids with a drag ratio of 2:1 (first row) and 100:1 (second row). Although the continuous optimal gaits are generated, the figure visualizes six optimal gaits. We set the color of each body trajectory to match the color of each shape trajectory. size of the maximum efficiency gait. The first row shows the shape trajectory of the Pareto-optimal steering gaits each system on $x$-CCF in isometric embedding coordinates. The second row shows the body trajectories of the swimmers generated by the optimal gait families. Based on the size of the gait cycle, the body trajectories also have different step sizes.}
	\label{fig:SteeringOptGaitResult}
\end{figure}

\cref{fig:SteeringOptGaitResult} presents Pareto-optimal steering gaits for three-link swimmers across varying forward-to-lateral drag ratios. As this ratio increases, the swimmer generates more forward motion due to an enhanced anchoring effect when pushing with a rotated link. The steering gait family transitions smoothly between forward and turning gaits, with greater turning motion observed when the gait center shifts toward the upper-right region of shape space.

The maximum forward-efficiency gaits produce pure forward motion, while maximum turning-efficiency gaits do not always yield pure rotation. For the viscous swimmer with a 2:1 drag ratio, the maximum-turning-efficiency gait induces slight backward motion. In contrast, for the perfect-fluid swimmer and the viscous swimmer with a 100:1 drag ratio, the maximum-turning gaits generate slight forward motion. Gaits unaffected by joint limits create arc-like trajectories, whereas those constrained by joint limits produce sharp turns due to the high concentration of $\theta$-CCF near these limits.

\begin{figure}
	\centering
	\includegraphics[width=0.8\columnwidth]{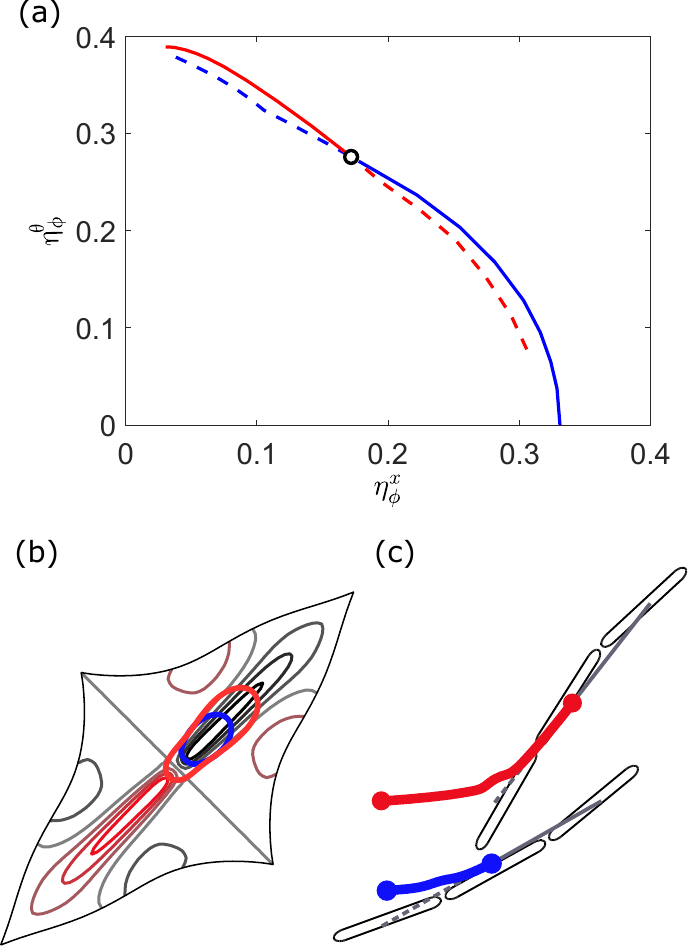}
	\caption{(a) Two branches of the Pareto front for a viscous three-link swimmer with a 100:1 drag ratio, shown in the forward–turning efficiency space. The red curve originates from the turning gait (turning-gait branch), and the blue curve originates from the forward gait (forward-gait branch). Solid lines indicate non-dominated fronts, while dashed lines indicate dominated fronts. The circle marks the intersection of the two branches (b) Two distinct gaits in the isometric embedding space, each corresponding to the intersection point of the two branches and yielding the same forward and turning efficiencies. The red gait produces a larger step size but at a higher cost, while the blue gait produces a smaller step size and lower cost. (c) The body trajectories generated by these two gaits. Since gait efficiency is defined as the exponential coordinate of net displacement $\gaitexpdisp$ divided by gait cost $\gaitcost$, both gaits achieve the same efficiency despite their differences in step size and cost.}
	\label{fig:Bifurcation}
\end{figure}

As illustrated in \cref{fig:Bifurcation}, bifurcations arise not only from inequality constraints but also from multiple solutions. Specifically, the viscous swimmer with a 100:1 drag ratio exhibits two Pareto front branches originating from forward and turning gaits, respectively. The turning-gait branch consists of larger-amplitude gaits than the forward-gait branch, leading to an intersection point. To the left of this point in $x$-$\theta$ efficiency space, the turning branch is slightly more Pareto-optimal than the forward branch, and vice versa. At the intersection, two distinct gaits achieve identical $x$- and $\theta$-efficiencies, but the turning-branch gait produces a larger step size while consuming more energy than the forward-branch gait.

\subsubsection{Baby-Step Gait Families}

To generate gait families with small step sizes, we formulate a family of optimization problems. Unlike the boundary problem—which seeks extremal gaits—we search for interior points along radial directions extending inward from the Pareto front. This problem is formulated as
\begin{equation}
    \constmintwo{\gaitparam}{\gaitcost}{\paretorad^2 = \contpar_1,}{\paretoangle = \contpar_2,}
    \label{eq:steeringbabygaitdef} 
\end{equation} 
where $\contpar_2$ is held constant throughout each optimization and is not treated as a continuation parameter.

\cref{fig:WholeGaitResults} illustrates the optimal steering gait family—including baby-step gaits—for three-link swimmers in both viscous and perfect fluids. To display evenly spaced gaits, we interpolate the gait parameters. As gaits approach the joint limit—when an inequality constraint transitions from inactive to active—a bifurcation occurs. Comparing the local and global approaches for the viscous swimmer reveals that the local method near these bifurcation points becomes unstable relative to those from the global method, likely due to the presence of multiple solutions.

For the viscous swimmer, we observe that the local search method can exhibit numerical instability under certain conditions:
\begin{enumerate}
	\item Joint Limit Contact: When a gait reaches the joint limit, the optimizer encounters a non-smooth region, causing bifurcation. This results in qualitative differences between neighboring gaits, leading to unevenly spaced points in \cref{fig:WholeGaitResults}. Drag-dominated systems are particularly susceptible to this issue.    
	\item Small Step Size: If a gait’s step size is too small, the local search method becomes highly sensitive to numerical techniques such as differentiation and error tolerance. This makes it difficult to construct an accurate predictor. In such cases, results from the global search method can be used instead.
\end{enumerate}

Note that although we generate the gaits inducing forward-left motion in this example, the gaits generating other motions such as backward-left or forward-right can be derived by a symmetry of CCF or executing the current gait in reverse.

%%%%%%%%%%%%%%%%%%%%%%%%% Conclusion %%%%%%%%%%%%%%%%%%%%%%%%%
\section{Conclusion}
In this paper, we established a mathematical framework for transitioning across optimal gaits within a family, thereby enhancing both controllability and maneuverability. Our primary contribution is a gait optimization framework that generates optimal gait families for locomoting systems in a single process, rather than optimizing each gait individually.

The global search approach constructs a reduced-order optimal gait family using a brute-force strategy, providing an overall intuition of the gait family's structure. The subsequent local search refines this family by optimizing higher-order gait parameters. Notably, the boundaries of the gait families consist of Pareto-optimal gaits that maximize forward movement efficiencies with turning. In contrast, constructing the interiors of the gait families parallels the process of identifying baby-step gaits with specified steering rates. We demonstrated these methods on three-link swimmers operating in both drag- and inertia-dominated environments. The resulting optimal gait families span regions of the $x$–$\theta$ efficiency space, and, unlike the linear boundaries observed in the average-velocity space of car-like robots, the boundaries for the locomoting systems studied here are nonlinear and feature nonsmooth points.

Looking ahead, our objective is to develop a computationally efficient, gait-based controller that leverages these gait families. The trajectory tracking problem for locomotion systems can be reduced to identifying an optimal sequence of gait families, which can be formulated as an optimal control problem for single rigid-body motion in either a continuous or hybrid (mixed discrete-continuous) domain \cite{frazzoli2001robust,kushleyev2009timebounded,flasskamp2019symmetry,bjelonic2022offline}. Additionally, effective gait switching and modulation are essential when the available gaits are not collocated. Large-amplitude transitions may force the robot to momentarily halt its current motion to reposition its joints before adopting a new gait—an abrupt change that can compromise overall smoothness and efficiency. We anticipate that these issues can be mitigated either by designing limit cycle attractors (or dynamic motion primitives) that track the optimal gait families \cite{hogan2013dynamic,ijspeert2013dynamical} or by integrating an optimal control strategy \cite{cabrera2024optimal,choi2024optimal}.

% \addtolength{\textheight}{-12cm}   % This command serves to balance the column lengths
% on the last page of the document manually. It shortens
% the textheight of the last page by a suitable amount.
% This command does not take effect until the next page
% so it should come on the page before the last. Make
% sure that you do not shorten the textheight too much.

\begin{appendices}
\appendices

\section{Higher-order Gait Parameter}
\label{app:gaitparam}
We use a truncated Fourier series as the gait parameters. With this parameterization, the $j$-th shape variable $\alpha_j(t)$ at time $t$ in the gait is calculated as
\begin{equation}
	\alpha_j(t) = a_{0,j} + \sum_{n=1}^k a_{n,j}\cos(n\omega t)+ b_{n,j}\sin(n\omega t),
\end{equation}
where $a_{n,j}$ is the $n$-th Fourier coefficient for $j$-th shape variable and $\omega$ is the frequency of the shape function and set as $2\pi$. Then, the gait parameter $p_j$ for $j$-th shape trajectory is expressed as:
\begin{equation}
	p_j = \transpose{\big[\overbrace{\begin{matrix}
			a_{0,j} & a_{1,j} & b_{1,j} & \cdots & a_{k,j} & b_{k,j} \\
		\end{matrix}}^{2k+1}\big]}.
\end{equation}

% \section{Gradient of the net motion}
% The gradient of the averaged body velocity, $\grad[\gaitparam] \gaitexpdisp$, represents the variation of the enclosed area with respect to the change of the gait cycle. The approximation in \eqref{eq:BVI} is a surface integral whose boundary is defined by the gait $\gait$; hence a variation around gait $\gait$ is intuitively thought of as adding or subtracting weighted regions near the surface boundary. Formally, the general form of the Leibniz rule converts this gradient of the functional with respect to variations of its boundary to the gradient of its boundary weighted by the integrand\cite{flanders1973differentiation}:
% \begin{equation}
% 	\grad[\gaitparam] \gaitexpdisp = \grad[\gaitparam] \iint_{\gait_a}{D(\mixedconn)} = \oint_\gait \left(\grad[\gaitparam] \gait \right) \intprod D(\mixedconn),
% 	\label{eq:gradBVI}
% \end{equation}
% where $\intprod$ is an interior product. The interior product contracts a differential 2-form (i.e., tensor field) $D(\mixedconn)$ with a vector field $\grad[\gaitparam] \gait$ to a 1-form (i.e., covector field).

\section{Gait Cost}
\label{app:gaitcost}
\begin{figure}
	\centering
	\includegraphics[width=0.5\textwidth]{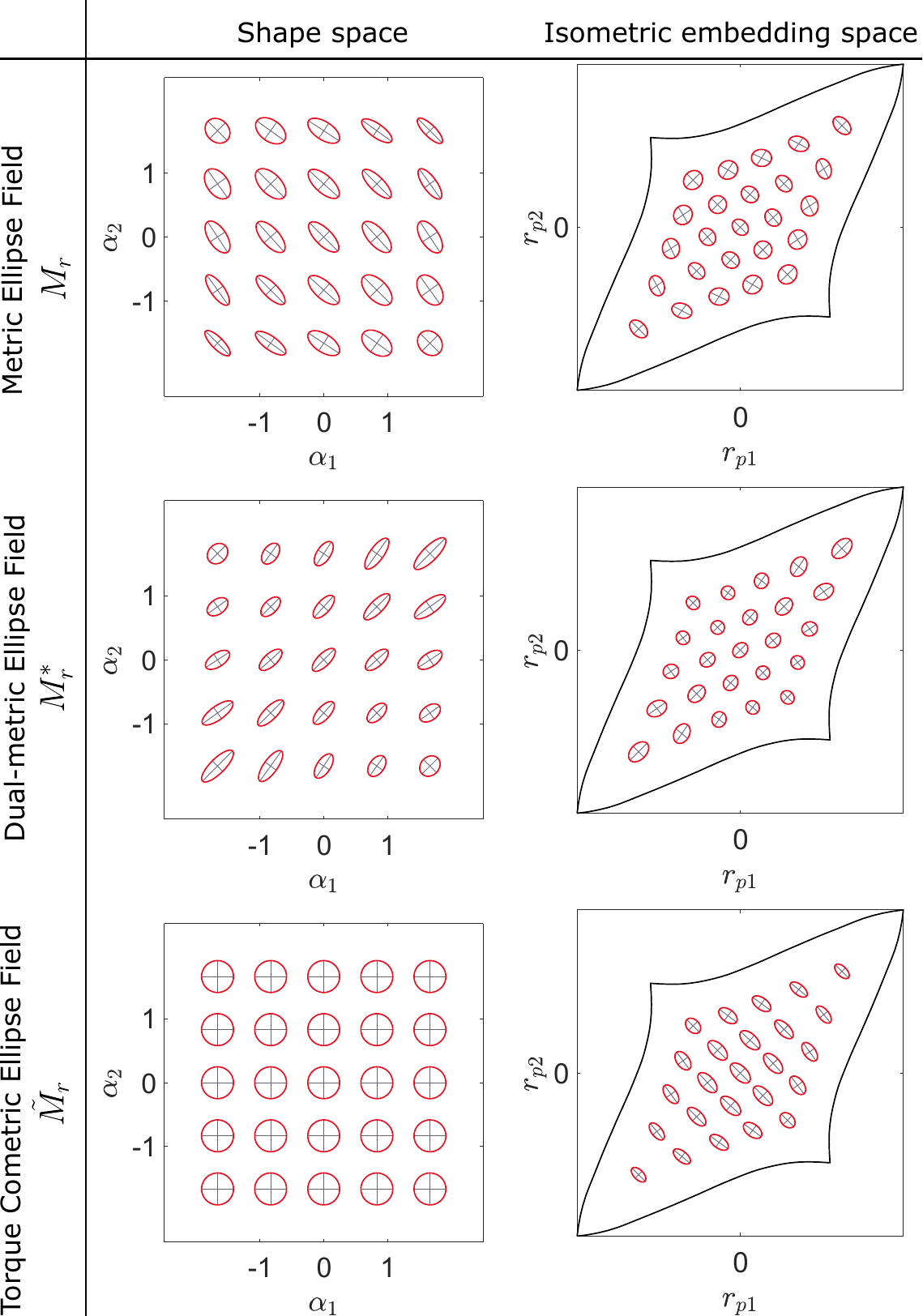}
	\caption{Tissot indicatrices for a variety of useful metrics on the perfect-fluid three-link swimmer \cite{cabrera2024optimal}. (Upper left) Infinitesimal circles on the manifold distort into ellipses in joint-angle space, indicating the relative effort for shape changes (called Tissot indicatrix). Longer axes denote more efficient directions. (Upper right) The isometric embedding minimizes distortion but retains ellipticality due to the manifold’s curvature. (Middle left) The dual metric $\dualmtensor$ illustrates how forces influence particle acceleration along the shape trajectory, with longer axes marking torque directions that are less effective. (Lower left) The actuation-based cometric $\comtensor$ more accurately captures actuator effort by indicating how forces act on the actuators. When the actuation configuration aligns with the joint configuration and all weights are equal, the Tissot indicatrices become circular. (Lower right) The actuation-based cometric tensor distorts into ellipses on the manifold under $\comtensor$.}
	\label{fig:MetricFields}
\end{figure}

\subsection{Drag-dominated Systems}
For drag-dominated systems, the metric tensor $\dtensor$ describes the effective drag coefficients as the systems change their shape in different directions. The drag tensor can be calculated by \cref{eq:metriccalc} with the local drag coefficient matrix on each link,
\begin{equation}
    \mu_i = \begin{bmatrix}
        L & &  \\
        & kL & \\
         & & \frac{1}{12}kL^3
    \end{bmatrix},
\end{equation}
where $k$ is the ratio between longitudinal and lateral drag coefficients. There exists the shape-dependent linear relationship between the torques $\tau$ on each joint and the generalized velocities,
\begin{equation}
	\tau = \dtensor\dot{r}.
	\label{eq:dissippower}
\end{equation}

The power dissipated by a drag environment is the inner product of joint velocities and torques, and it is represented by the metric norm of the velocity. Hence, the effort to execute the gait is identical to the weighted-perimeter $\alnth_\gait$, which can be calculated by the path integral along the gait on the manifold associated with the drag tensor $\dtensor$:
\begin{equation}
    \alnth_{\gait} = \oint_{\gait} \norm{\dot{r}}_{\dtensor}dt.
    \label{eq:dragcostriem}
\end{equation}

% The gradient of the cost measures how variations of gait affect its execution cost per cycle. The gradient component of this term can be derived as the gradient of the arclength on metric-normalized space:
% \begin{equation}
% \grad[\gaitparam] \alnth_\gait = \oint_{\gait}{\frac{1}{2\norm{\dot{r}}_{\mtensor}}\left(2\innerprod{\grad[\gaitparam]\dot{r}}{\dot{r}}_{\mtensor} + \innerprod{\dot{r}}{\dot{r}}_{\left(\grad[\gaitparam]\mtensor\right)}\right)}dt.
% \label{eq:dragcostgrad}
% \end{equation}

\subsection{Inertia-dominated Systems}
For inertia-dominated systems, the metric tensor $\mtensor$ describes the system's inertial information with respect to the shape variables~\cite{hatton2013geometric}. The acceleration-based cost considers the force norm with the dual metric, $\dualmtensor$, which is chosen by the inverse of the mass metric. Thus, the dual metric ellipses in \cref{fig:MetricFields} are orthogonal to the metric ellipses. This cost measures how much the gait generates the covariant acceleration norm:
\begin{equation}
	\norm{\tau}_{\dualmtensor}^2 = \norm{\covacc}_{\mtensor}^2
	\label{eq:acccostdef}
\end{equation}

The actuation-based cost considers the force norm with the cometric tensor, $\comtensor$, describing the physical configuration of the actuator~\cite{cabrera2024optimal}. For the three-link swimmer, we assume that each actuator is attached to each joint and associates the same ratio of efforts. Then, as illustrated in \cref{fig:MetricFields}, the cometric reduces to the identity matrix. The actuation cost functional is defined as the squared norm of the actuator torque:
\begin{equation}
	\norm{\tau}_{\comtensor}^2 = \norm{\tau}^2
	\label{eq:actcostdef}
\end{equation}
In particular, the cost to execute the gait is associated with the notion of how long it take to execute gait cycle with an unit-effort shape velocity. Then, the gait cost is equal to the fourth root of the gait cost with unit period,
\begin{equation}
	\gaitcost = \left(\int^1_0\norm{\tau}^2dt\right)^{1/4}
\end{equation}

% The gradient of the cost is defined as
% \begin{equation}
% 	\grad[\gaitparam]{\gaitcost} = \frac{1}{4\gaitcost^3}\int^1_0\left(2\tau\grad[\gaitparam]{\tau}\right)dt
% \end{equation}

\subsection{Isometric embedding space}
To visualize the metric-weighted space (called the isometric embedding space), the previous work \cite{hatton2017kinematic} defines the indicatrix ellipse. As shown in the first column of \cref{fig:MetricFields}, the indicatrix ellipse describes the scale and direction of the distortions. A Tissot transformation matrix $T$ can be derived from the singular value decomposition of the metric tensor $\mtensor$:
\begin{equation}        
    T = U\Sigma^{-\frac{1}{2}}\transpose{U} \ \text{ for } \  \svd(\mtensor) = U\Sigma\transpose{U}.
\end{equation}
Here, the columns of $U$ represent the direction of each axis of the ellipse, and the diagonal elements of $\Sigma$ represent the singular values which is the magnitude of each axis.

As shown in the first row of \cref{fig:MetricFields}, the isometric embedding space is defined by stretching the shape space to minimize the distortion of the indicatrix ellipse at every point (i.e., the shape of the indicatrix ellipse is close to the circle in this space.) The actual cost for the drag-dominated system can be approximated by the path length in the isometric embedding space. For the inertial-dominated system, the acceleration-based cost is the covariant acceleration norm in the isometric embedding space.

There must be a map from the shape space to the isometric embedding space:
\begin{equation}
	\Psi : r \mapsto r_p,
\end{equation}
where $r$ is the old coordinate corresponding to the shape variables, and $r_p$ is the new coordinate.

\section{Mathematical Optimization}
\label{app:optimization}
Here, we review basic notions and terms for mathematical optimization problems in the paper. 

The family of optimization problems to construct the system of nonlinear equations is formulated as 
\begin{equation}
	\constmintwo{\gaitparam}
	{\objfun(\gaitparam,\contpar)}
	{\eqfun[i](\gaitparam,\contpar) = 0, \ i \in \{1,\cdots,\numeq\}}
	{\ineqfun[i](\gaitparam,\contpar) \leqslant 0, \ i \in \{1,\cdots,\numineq\}}
	\label{eq:convoptprob}
\end{equation}
where $\gaitparam \in \mathcal{P} \coloneqq \euclid^m$ is optimization variables, $\numeq$ is the number of equality constraints, and $\numineq$ is the number of inequality constraints, $\objfun:\mathcal{P}\times \mathcal{C}\to\euclid$, $\eqfun[i]:\mathcal{P}\times \mathcal{C} \to \euclid$, and $\ineqfun[i]:\mathcal{P}\times \mathcal{C} \to \euclid$ are the objective function, the equality constraint function, and the inequality constraint function, respectively. 

The KKT conditions are first-order optimality necessary conditions for a solution in nonlinear optimization problems and the generalization of the method of Lagrange multipliers to include inequality constraints.
\begin{definition}
Every optimal solution $(\gaitparam^*,\objmult^*,\eqmult^*,\ineqmult^*,\contpar)$ to the optimization problem in \eqref{eq:convoptprob} must satisfy below conditions:
\begin{itemize}
    \item Stationarity condition:
    \begin{subequations}
        \begin{equation}
            \grad[\gaitparam] \lagrfun = 
            \objmult\grad[\gaitparam] \objfun +
            \sum_{j=1}^{\numeq}\eqmult[j]\grad[\gaitparam] \eqfun[j] + 
            \sum_{k=1}^{\numineq}\ineqmult[k]\grad[\gaitparam] \ineqfun[k] = \mathbf{0}, 
        \end{equation}
        for
        \begin{equation}
            \objmult^2 + \norm{\eqmult}^2 + \norm{\ineqmult}^2 = 1.				
        \end{equation}
    \end{subequations}
    $\objmult$ is the scalar parameter, and 
    \begin{subequations}
        \begin{gather}
            \objmult > 0, \text{ for minimizing } \objfun(\gaitparam^*,\contpar), \\
            \objmult < 0, \text{ for maximizing } \objfun(\gaitparam^*,\contpar).
        \end{gather}
    \end{subequations}
    \item Primal feasibility:	
    \begin{subequations}
        \begin{gather}
            \eqfun(\gaitparam^*,\contpar) = \mathbf{0}, \\
            \ineqfun(\gaitparam^*,\contpar) \preceq \mathbf{0}.
        \end{gather}
        \label{eq:kktprimfeas}
    \end{subequations}
    \item Dual feasibility:
    \begin{equation}
        \ineqmult \succeq \mathbf{0},
        \label{eq:kktdualfeas}
    \end{equation}
    \item Complementary slackness:
    \begin{equation}
		\sum_{k=1}^{\numineq}\ineqmult[k]\cdot\ineqfun[k](\gaitparam^*,\contpar) = 0,
        \label{eq:kktcompslack}
    \end{equation}
\end{itemize}
\end{definition}
where $\preceq$ denotes that every element of the vector on the left-hand side is less than those of the right-hand side. In the same token, $\succeq$ is the opposite of $\preceq$. Because of complementary slackness, the inequality constraint can be active or inactive. If $\ineqfun[i](\gaitparam^*,\contpar) = 0$, the inequality constraint becomes active, and the $\ineqmult[i]$ can be nonzero. If not, it becomes inactive, and $\ineqmult[i]$ should be zero.

\begin{definition}[Feasible Point]
    $\gaitparam$ is called the \emph{feasible point}, if all constraints are satisfied at $\gaitparam$ (i.e.,\ it is on the domain of the cost and constraint functions.) A \emph{feasible set} or \emph{feasible region} is the set of all feasible points.
\end{definition}

Note that the KKT conditions provide the necessary conditions for optimality. To verify sufficiency, one can apply the second-order sufficient conditions.
\begin{definition}[Sufficient Condition]
	\label{thm:secondsuff}
	$(\gaitparam^*,\objmult^*,\eqmult^*,\ineqmult^*)$ is a strict local minimizer if the point satisfies KKT conditions, and
	\begin{equation}
		\transpose{d}\nabla_{\gaitparam}^2\lagrfun d > 0,
	\end{equation}
	for all $d \in \euclid^{\numeq+m_{\text{active}}}$ such that
	\begin{equation}
		\transpose{\begin{bmatrix}
			\grad[\gaitparam]\eqfun & \grad[\gaitparam]\ineqfun[\text{active}]
		\end{bmatrix}} d = \mathbf{0},
	\end{equation}
	where $\ineqfun[\text{active}]$ is the vector-valued function to collect active inequality constraints and $m_{\text{active}}$ is the number of active inequality constraints.
\end{definition}

\section{Joint limit as Inequality Constraint}
\label{app:jointlimit}
Let discretize the $j$-th shape variable $\alpha_j$ by the time $t_k \in [0,T]$ evenly. Let $b_\text{ub}$ and $b_\text{lb}$ be the upper and lower bound of the shape variables, respectively. Then, the joint limit can be expressed as 
\begin{subequations}	
	\begin{gather}
		\alpha_j(t_k) \leqslant b_\text{ub} \\ 
		\alpha_j(t_k) \geqslant b_\text{lb}
	\end{gather}
	\label{eq:shapeineqcond}
\end{subequations}
If the time $t_k$ is constant, the Fourier coefficients $A_k$ becomes the constant value. For the constant Fourier coefficients, the shape variables can be expressed as
\begin{equation}
	\alpha_j(t_k) = A_kp_j,
	\label{eq:fourtranswaypoint}
\end{equation}
where
\begin{equation}
	A_k = \begin{bmatrix}
		1 & \cos(\omega t_k) & \sin(\omega t_k) &  \cos(2\omega t_k) & \cdots
	\end{bmatrix}.
\end{equation}
We can formulate the inequality constraint by combining 
\eqref{eq:shapeineqcond} with \eqref{eq:fourtranswaypoint}:
\begin{equation}
	\ineqfun[i](\gaitparam,\contpar) =
	\begin{cases}
		A_k p_j - b_\text{ub}, & \text{for upper bound}, \\
		-A_k p_j + b_\text{lb}, & \text{for lower bound}.
	\end{cases} 
	\label{eq:ineqcond}
\end{equation}
If we gather the set of \eqref{eq:ineqcond} by distributing the evenly-spaced time $t_k$ and order them properly, the joint limit becomes the set of linear inequality constraints.

\section{Numerical Continuation}
\label{app:numcont}
\subsection{Handling Inequality}
In the presence of the inequality constraint, we need to consider the primal, dual feasibility, and complementary slackness. This problem is another form of the nonlinear complementary problem (NCP). That makes it difficult to formulate the set of nonlinear equations $\contfcn(q,\contpar)$. To formulate this condition, the optimization community has used the NCP function \cite{fischer1992special,fischer1997solution}:
\begin{equation}
	\ncpfun(a_i,b_i) \coloneqq \sqrt{a_i^2+b_i^2}-a_i-b_i.
\end{equation}
It allows these conditions to be expressed as the nonsmooth system of equations because 
\begin{equation}
	\ncpfun(a_i,b_i) = 0 \Longleftrightarrow a_i,b_i \geqslant 0, \ a_i \text{ or } b_i = 0.
\end{equation}
Thus, if we replace $a_i$ and $b_i$ as $\ineqmult[i]$ and $-\ineqfun[i]$, it forces the point to satisfy the feasibility and complementary conditions for inequality constraint.

To avoid the bifurcation caused by the inequality constraints, we push the point into the feasible region. In our problem, the inequality constraint is expressed as the form of NCP functions. They can be singular if both inputs ($\ineqmult[i]$ and $\ineqfun[i]$) are zeros. To overcome this non-smoothness and predict the next point in the feasible space, we adjust the predicted point so that
\begin{equation}
	\ineqmult[i] \leftarrow \begin{cases}
		0 & \text{for } i \in \{ i \ | \ \ineqfun[i] < -\varepsilon \}\\
		0  & \text{for } i \in \{ i \ | \ \ineqmult[i] < 0 \}\\
		\delta  & \text{for } i \in \{ i \ | \ \lvert\ineqfun[i]\rvert < \varepsilon \}
	\end{cases},
	\label{eq:corrjumpstep}
\end{equation}
where $\varepsilon$ is the tolerance for checking the violence of the inequality constraints, and $\varepsilon > 0$. Each case in \eqref{eq:corrjumpstep} corresponds to the situation to violate the primal feasibility, dual feasibility, and complementary slackness in order. The purpose of this step is to jump over the nonsmooth region of NCP functions. Even if the jump in \eqref{eq:corrjumpstep} only moves the inequality dual variables to the feasible region, the corrector step can adjust other elements so that they satisfy the KKT conditions.

% \subsection{Jacobian of Continuation Function}
% Let define $\contsol = \transpose{[\contvar, \contpar]}$. Then, the Jacobian of the continuation function is 
% \begin{equation}
% 	\transpose{\grad[\contsol] \contfcn} = \begin{bmatrix}
% 		\grad[\gaitparam]^2 \lagrfun & \grad[p]\objfun & \grad[p]\eqfun & \grad[p]\ineqfun & \mathbf{0} \\
% 		\mathbf{0} & 2\objmult & 2\transpose{\eqmult} & 2\transpose{\ineqmult} & 0 \\
% 		\grad[\gaitparam]{\transpose{\ncpfun_1}} & 0 & \transpose{\mathbf{0}} & \begin{matrix} \parderiv{\ncpfun_1}{\ineqmult[1]} & \cdots & 0 \end{matrix} & 0 \\
% 		\vdots & \vdots & \vdots & \ddots & \vdots \\		
% 		\grad[\gaitparam]{\transpose{\ncpfun_{\numineq}}} & 0 & \transpose{\mathbf{0}} & \begin{matrix} 0 & \cdots & \parderiv{\ncpfun_{\numineq}}{\ineqmult[\numineq]} \end{matrix} & 0 \\
% 		\transpose{\grad[p]\eqfun} & 0 & \transpose{\mathbf{0}} & \transpose{\mathbf{0}} & \grad[c]\eqfun
% 	\end{bmatrix}
% \end{equation}
% for
% \begin{subequations}
% 	\begin{gather}
% 		\ncpfun_{i} = \ncpfun(\ineqmult[i],-\ineqfun[i]) \\
% 		\parderiv{\ncpfun_i}{\ineqmult[i]} = 
% 		\frac{\ineqmult[i]}{\sqrt{\ineqmult[i]^2+\ineqfun[i]^2}}-1, \\ 	
% 		\grad[\gaitparam]{\ncpfun_{i}} = 
% 		\left(\frac{\ineqfun[i]}{\sqrt{\ineqmult[i]^2+\ineqfun[i]^2}}+1\right) \grad[\gaitparam] \ineqfun[i].
% 	\end{gather}	
% \end{subequations}
% Note that the derivative of NCP functions is not defined when both $\ineqmult[i]$ and $\ineqfun[i]$ are zeros.

\subsection{Algorithm}
\begin{algorithm}
\caption{Predictor-Corrector Method}\label{alg:pc}
\begin{algorithmic}
\Require $\gaitparam^{(0)}, \contpar^{(0)}$
\State Find $\objmult^{(0)},\eqmult^{(0)},\ineqmult^{(0)}$ to satisfy KKT conditions.
\State $\contsol^{(0)} \gets [\gaitparam^{(0)},\objmult^{(0)},\eqmult^{(0)},\ineqmult^{(0)}, \contpar^{(0)}]$
\State $i \gets 0$
\Repeat
\State $\contsearchvec \gets $ searching directions.
\State $\contpredstep \gets \text{Proj}_{\nullcont}\left(\contsearchvec\right).$ \Comment{Compute tangent vector}
\State $v \gets \contsol^{(i)} + h\contpredstep$
\Repeat \Comment{Newton Raphson Method}
\State $s \gets \dotprod{\contpredstep}{(v-\contsol^{(i)})}$ \Comment{Pseudo step length}
\State $w \gets v - \pinv{\begin{bmatrix}    \nabla\contfcn \\ \transpose{\contpredstep}\end{bmatrix}}\begin{bmatrix}\contfcn \\ s - h\end{bmatrix}$
\State
\State Adjust $\ineqmult[i]$ based on \cref{eq:corrjumpstep}
\State $v \gets w$
\Until $v$ converges
\State $i \gets i + 1$
\State $\gamma^{(i)} \gets v$
\Until satisfying stop conditions
\end{algorithmic}
\end{algorithm}

\section{Local Bifurcation}
\label{app:bifurcation}
Here, we review basic notions and terms for bifurcation in the paper. We are specifically interested in a local bifurcation that occurs at the singularity. The detail can be found in \cite{guckenheimer2013nonlinear, allgower2003introduction}. Consider the system of ordinary differential equations (or the vector field)
\begin{equation}
	\frac{d\contvar}{dt} = \contfcn(\contvar(t),\contpar(t)).
\end{equation}
If we consider the evolution of the systems under varying $\contpar$, there may be a point that remains stationary over time, and it is called a ``stationary point''. In this paper, we use the numerical continuation method to find stationary points of $\contfcn$ (i.e., at the stationary point, the optimality necessary condition is satisfied.) If the optimality sufficient condition is satisfied at the stationary point, the point is one of the local optima.
\begin{definition}[Stationary points]
	A point $(\contvar,\contpar)$ is called a \emph{stationary point} of $\contfcn$ if
	\begin{equation}
		\contfcn(\contvar,\contpar) = \mathbf{0}.
	\end{equation}
\end{definition}
A stationary point often be substituted by ``equilibrium solution'', ``fixed point,'' or ``critical point.'' 

\begin{definition}[Regular points]
	A stationary point $(\contvar,\contpar)$ is called a \emph{regular point} of $\contfcn$ if the derivative of $\contfcn$ has a full rank. The stationary point is called a \emph{singular point} if it is not regular.
\end{definition}

\begin{theorem}[Implicit function theorem]
	Let $\contfcn : \euclid^{n+k} \to \euclid^n$ be a continuously differentiable function, $\contvar \in \euclid^n$, and $\contpar \in \euclid^k$. There exists $\contvar(\contpar)$ so that $(\contvar,\contpar)$ is always a stationary point if the Jacobian of $\contfcn$ has a full rank.
\end{theorem}

The implicit function theorem implies that the optimal gait family (i.e., stationary points) can be expressed as the smooth functions of $\contpar$ away singular points. At singular points, the optimal gait families may split into several branches, and the points are called \emph{bifurcation points}. At the bifurcation point, the sudden qualitative change of the system could happen by a small smooth change of $\contpar$. As the term ``bifurcation'' is general and includes vagueness, we focus attention on the local bifurcation which could be analyzed by the stability of a stationary point in terms of a gait rather than gait parameters.
\end{appendices}

%\section*{ACKNOWLEDGMENT}
%We thank the National Science Foundation for support via Grant CMMI-1653220 and the Office of Naval Research via Award N00014-23-1-2171.

%%%%%%%%%%%%%%%%%%%%%%%%%%%%%%%%%%%%%%%%%%%%%%%%%%%%%%%%%%%%%%%%%%%%%%%%%%%%%%%%

\bibliographystyle{IEEEtran}
\bibliography{ref}

\end{document}